\documentclass[12pt,a4paper]{article}

\usepackage{charter}
\usepackage[left=3.5cm, right=2.5cm]{geometry}
\usepackage{amssymb}
\usepackage[smaller,printonlyused,withpage]{acronym}
\usepackage[pdftex]{graphicx} \graphicspath{{./images/}}
\usepackage[below]{placeins}
\usepackage[usenames]{color}
\usepackage[polish,english]{babel}
\usepackage[T1]{fontenc}
\usepackage[utf8]{inputenc}
\usepackage{arabtex}

\usepackage{amsmath}
\usepackage{listings}
\usepackage{enumerate}
\usepackage{multirow}
\usepackage{multicol}
\usepackage{array}
\usepackage{verbatim}
\usepackage{fancyhdr}
\usepackage{float}
\floatstyle{plaintop}
\restylefloat{table}
\usepackage[font=small,labelfont=bf]{caption}
\usepackage{eurosym}
\usepackage[final]{pdfpages}
\usepackage{adjustbox}
\usepackage{blindtext}

\usepackage[colorinlistoftodos]{todonotes}

\usepackage{makeidx}

\usepackage{colortbl}
\usepackage{booktabs}
\usepackage{multirow}
\usepackage{makecell}
\usepackage{subcaption}
\usepackage[table,xcdraw]{}
\usepackage{authblk}
\makeindex

\usepackage[pdftex,bookmarks,pdfpagelabels]{hyperref}

\usepackage{listings}
\begin{document}
\author[+]{Ludwik Bukowski\thanks{the Author of this thesis and corresponding author}}
\author[+]{Witold Dzwinel\thanks{MsC thesis supervisor}}
\affil[+]{AGH University of Science and Technology, WIET, Department of Computer Science, Kraków, Poland}
\title{SuperNet - An efficient method of neural networks ensembling}
\maketitle

\begin{abstract}

The main flaw of neural network ensembling is that it is exceptionally demanding
computationally, especially, if the individual sub-models are large neural networks,
which must be trained separately. Having in mind that modern DNNs can be very
accurate, they are already the huge ensembles of simple classifiers, and that one can
construct more thrifty compressed neural net of a similar performance for any
ensemble, the idea of designing the expensive SuperNets can be questionable. The
widespread belief that ensembling increases the prediction time, makes it not attractive
and can be the reason that the main stream of ML research is directed towards
developing better loss functions and learning strategies for more advanced and efficient
neural networks. On the other hand, all these factors make the architectures more
complex what may lead to overfitting and high computational complexity, that is, to the
same flaws for which the highly parametrized SuperNets ensembles are blamed. The
goal of the master thesis is to speed up the execution time required for ensemble
generation. Instead of training K inaccurate sub-models, each of them can represent
various phases of training (representing various local minima of the loss function) of a
single DNN [Huang et al., 2017; Gripov et al., 2018]. Thus, the computational
performance of the SuperNet can be comparable to the maximum CPU time spent on
training its single sub-model, plus usually much shorter CPU time required for training
the SuperNet coupling factors.
\end{abstract}

\newcommand{\ackname}{Acknowledgements}


\pagenumbering{roman}
\section*{\ackname}
\textit{We would like to thank:} 
\begin{enumerate}
\item \textit{A. Kolonko, P.Wandzel and K.Zuchniak}, for their work
resulting in promising preliminary results of \textit{SuperNet} design which stimulate this thesis.
\item \textit{ACK Cyfronet i PlGrid}, for providing computing resources.
\end{enumerate}

The work was partially supported by National Science Centre Poland NCN: \textit{project Number 2016/21/B/ST6/01539}.

\tableofcontents

\listoftables
\addcontentsline{toc}{section}{\listtablename}
\listoffigures
\addcontentsline{toc}{section}{\listfigurename}

\clearpage

\pagenumbering{arabic}
\section{Introduction} 
\subsection{Motivation and Thesis Statement}

One can say that there is a trade-off between the computational budget used for building neural networks and its final performance. Having in mind classification problems, simple models achieve lower accuracy while the complex ones require more CPU for training. 
Given the above, an interesting characteristic turn out to have ensembles of pre-trained networks. They tend to achieve better results than single models\cite{ensemble_good}, and the gain increases, especially, when the base models are significantly different. However, we still need additional resources to train the base models(in this dissertation we will use the term: the sub-models). The interesting question for that problem is whether we can have higher accuracy using ensembles, compared to single networks, using the same amount of computational budget.
Inspired by the most recent researches around machine learning, we were exploring the method of stacked ensembles of neural networks. We achieve the ensemble by merging multiple sub-networks with the last layer, which we additionally train.
We expect that such the architecture(we call it the \textit{Supermodel} or \textit{SuperNet}) can improve prediction accuracy because the output of neural networks is the most sensitive on the changes of weights from the last layer (see e.g., vanishing gradient problem).
It can be time efficient as well, assuming that we can train sub-models concurrently or obtain them as snapshots during the training of a single network\cite{m_free}, \cite{garipov}. The very preliminary results of such an approach were initially presented at the V International AMMCS Interdisciplinary Conference, 2019\cite{kolonko}. 
Mainly, the primary motivation of my thesis was to: 
\begin{enumerate}
\item Prove that \textit{Supermodeling} can be an efficient method for the accuracy improvement of neural networks.
\item Show the efficiency of \textit{Supermodeling} method, concerning the CPU time.
\item compare \textit{Supermodels} to the classical networks of the same size.
\item Investigate the method of obtaining sub-models for \textit{SuperNet} as snapshots of training a single model.
\end{enumerate}

\subsection{Structure of the thesis}

The first chapter gives a brief introduction to the theoretical background and the latest achievements in the area of convolutional neural networks.  We describe the datasets used and the technical aspects of our research. In the second chapter, we present a simple and efficient method for quick accuracy improvement of deep neural networks. We measure the properties of the ensembles built from partitioned fully-connected networks. In the third chapter, we present the results of \textit{Supermodeling} for deeper architectures. In the fourth chapter, we show that ensembles may be time-efficient when we consider novel methods for obtaining submodels. The final chapter contains conclusions and the future work around this subject.

\subsection{Contributions}

The contributions of the thesis are as follows:
\begin{enumerate}
\item We present novel method for quick accuracy gain for the neural networks by training only its last layer only. We present the results for a few different architectures and datasets.
\item We measure the performance of a partitioned network in comparison to a single model. We confirm the hypothesis that we can profitably replace very shallow dense networks with ensembles of the same size.
\item We show that \textit{Supermodels} of neural networks improve the accuracy. We achieve 92.48\% validation accuracy on the \textit{CIFAR10} dataset and 73.8\% on \textit{CIFAR100} which reproduce the best scores for the year 2015\cite{benchmarks}.
\item We compare the accuracy of the \textit{Supermodel} build from \textit{a)} snapshots of training a single network and \textit{b)} the independently trained models. We present that such an approach can be efficient for deeper architectures.
\end{enumerate}

\subsection{Research Challenges}

In my thesis, we are facing all the challenges related to the nature of classification problems. Reliable studies in that branch require not only the basics of machine learning but the knowledge of the newest achievements as well. 
The rapid increase of the number of papers published in this domain shows its top importance for development of artificial intelligence tools. Many of publications have very empirical character. The lack of a formalized way of defining introduced solutions raises many concerns.
Considering the more technical aspect of our work, the training of deep models is time-consuming, which takes even more resources when we consider hyperparameters's tuning and experiments on different datasets. 

\section{Technological Background and Related Work} 
\label{chap:background-related-work}

\subsection{Machine Learning and Pattern Recognition}

In recent years, machine learning is the most rapidly growing domain of computer science\cite{analyzed}. Neural networks prevail researchers' attention due to new methods of training and new type of architectures (deep, recurrent etc). Image recognition is the field, in which lead convolutional neural networks. This trend started to occur in 2012 with ImageNet-2012 competition's winner, the \textit{AlexNet}\cite{alexnet}, which showed the benefits of convolutional neural networks and backed them up with record-breaking performance\cite{alex_result}. Inspired by that spectacular success, the next years resulted in many architectures that focused on increasing accuracy even more\cite{vgg},\cite{going_deeper}, \cite{rethinking}, \cite{deep_res_learning}.
The most straightforward solution for improvement was to add more layers to the model\cite{vgg}, \cite{going_deeper}. However, the multiple researches had proven \cite{deep_res_learning},\cite{train_very_deep},\cite{const_time} that there are limitations of such method and very deeper networks lead to higher error. One of the problems was that neural network's accuracy decreases over many layers due to \textit{vanishing gradient problem}; as layers went deep(i.e., from output to input due to backpropagation learning scheme), gradients got small, leading to degradation. The interesting solution is proposed in research "Deep Residual Learning for Image Recognition"\cite{deep_res_learning}. The Authors introduced residual connections which are essentially additional connections between non-consecutive layers. That simple concept significantly improves the back-propagation process thus reduces training error and consequently allows to train deep networks more efficiently. Moreover, a similar approach introduces "Highway Networks"\cite{highway_networks} paper, in which the Authors accomplished a similar goal with the use of learned gating mechanism inspired by Long Short Term Memory(LSTM) recurrent neural networks\cite{lstm}. 

There is another important problem related to deep networks, called \textit{diminishing feature reuse} problem, which was not solved in the described architectures. Features computed by first layers are washed out by the time they reach the final layers by the many weight multiplications in between. Additionally, more layers results with longer training process. To tackle those complications, "Wide Residual Networks"\cite{wide_res_nets} were developed.
The original idea behind that approach was to extend the number of parameters in a layer and keep the network's depth relatively shallow. Having that concept in mind, together with some additional optimizations (e.g., \textit{dropout}\cite{dropout}) the Authors achieved new state-of-the-art results on accessible benchmark datasets\cite{benchmarks}.
In the contrast to the idea of dedicated, residual connections, the concept of "FractalNet"\cite{fractalnet} was developed. The Authors tackle the problem of vanishing gradients with an architecture that consists of multiple sub-paths, with different lengths. During the fitting phase, those paths are being randomly disabled, which adds the regularization aspect to the training. Their experiments reveal that very deep networks (with more than 40 layers) are much less efficient than their fractal equivalents.
The concept of disabling layers exploited "Stochastic depth"\cite{stochastic} algorithm. The method aims to shrink the depth of the network during training while keeping it unchanged in the testing phase. That was achieved simply by randomly skipping some layers entirely.
Going through the literature, one can observe that such concepts are often interpreted as specific variations around network ensembles\cite{res_as_ensemble}. 
In the next paragraph, we would like to focus on the concept of models ensemble.


\subsection{Ensemble and Stacked Generalization}

Ensemble learning is the machine learning paradigm of combing multiple learners together in order to achieve better prediction capabilities than constituent models\cite{ensemble_good}. The technique address the "bias-variance" dilemma\cite{bias_variance_dilema} in which such an ensemble can absorb too high bias or variance of the predictions of its sub-models and result in stronger predictor. The common types of ensembles are Bagging\cite{bagging}, Boosting\cite{boosting_alg} and Stacking. 
The underlying mechanism behind Bagging is the decision fusion of its submodels. The boosting approach takes advantage of the sequential usage of the networks to improve accuracy continuously.
In our work, we are focusing on the last method - Stacking.
Initially, the idea was proposed in 1992 in the paper "Stacked Generalization"\cite{stacked_generalization}. Conceptually, The approach is to build new meta learner that learns how to best combine the outputs from two or more models trained to solve the same prediction problem. Firstly, we need to split the dataset into train and validation subsets.  We independently fit each contributing model with the first dataset, called \textit{level 0} data. 
Then, we generate predictions by passing validation set through base models. Those outputs, named \textit{level 1} data, we finally use for training the meta learner. Authors of \cite{stacked_regression} paper improved the idea with a k-fold cross-validation technique for obtaining sub-models. In \cite{super_learner} it has been demonstrated that such a supermodel results in better predictive performance than any single contributing submodel.
Base submodels must produce uncorrelated predictions. The NN Stacking works best when the models that are combined are all \textit{skillful}, but \textit{skillful} in different ways. We can achieve it by using different algorithms. 

This brief introduction should help to understand what is the state-of-the-art and the current challenges when building highly accurate and deep convolutional neural networks(CNNs). In my thesis, we propose the ensemble architecture, which we are going to describe and analyze in the following chapters.

\subsection{Data and technical aspects}

For all of our experiments, we used the publicly available datasets that are commonly used as reference sets in the literature. It seems natural to use well-known data so that the results could be easily replicated or compared with different analyses. The datasets we are referring to are shwon in Table \ref{tab:datasets}. 
The Table \ref{tab:mymodel} presents models used for experiments related to \textit{CIFAR10} dataset. Moreover, in the thesis we present the results on \textit{DenseNet}\cite{densenet}, \textit{VGG-16}\cite{vgg} and multilayer perceptrons(MLP).
In our experiments, we are training a network on the training set and validate the \textit{accuracy} and \textit{loss} on the validation set. For some experiments, we isolated the testing set, which we used for final model evaluation. On most charts, we present the validation accuracy and loss, as we regard them as more important than the training curves. Training accuracy and loss always show an improvement, which still may be overfitting though. 
All the code was written in \textit{Python3} and architectures modeled in \textit{Keras}\cite{keras} library.
The code was run on \textit{GPGPU} on the supercomputer \textit{Prometheus}\cite{prometheus}. The CPU measured was the value returned of the field \textit{CPUTime} of \textit{sacct} command run on \textit{Prometheus} for the particular job.

\begin{table}[]
\centering
\begin{tabular}{|l|l|l|l|}
\hline
                       & \textit{samples} & \textit{input vector}                                              & \textit{classes} \\ \hline
\textit{CIFAR10}       & 60,000                     & 32x32x3                                                            & 10                         \\ \hline
\textit{CIFAR100}      & 60,000                     & 32x32x3                                                            & 100                        \\ \hline
\textit{Fashion MNIST} & 60,000                     & 28x28                                                              & 10                         \\ \hline
\textit{Newsgroups20}  & 20,000                     & \begin{tabular}[c]{@{}l@{}}text, \\ different lengths\end{tabular} & 20                         \\ \hline
\end{tabular}
\caption[Datasets]{Datasets used in my thesis.}
\label{tab:datasets}
\end{table}

\begin{table}[]
\centering
\begin{adjustbox}{width=0.9 \textwidth}
\begin{tabular}{@{}|l|l|l|@{}}
\toprule
\textit{Layer}           & \textit{Parameters}       & \textit{Details}                                             \\ \midrule
Input           & 32 x 32 x 3      & CIFAR10 input image                                 \\ \midrule
Convolutional   & 32 (3x3) filters & activation elu, batch norm, L2 reg  $\alpha$=0.0001 \\ \midrule
Convolutional   & 32 (3x3) filters & activation elu, batch norm, L2 reg $\alpha$=0.0001 \\ \midrule
MaxPooling      & 2x2              & 02. dropout                                         \\ \midrule
Convolutional   & 64 (3x3) filters        & activation elu, batch norm, L2 reg $\alpha$=0.0001\\ \midrule
Convolutional   & 64 (3x3)         & activation elu, batch norm, L2 reg $\alpha$=0.0001\\ \midrule
MaxPooling      & 2x2              & 0.3 dropout                                         \\ \midrule
Convolutional   & 128 (3x3) filters        & activation elu, batch norm,L2 reg $\alpha$=0.0001\\ \midrule
Convolutional   & 128 (3x3)        & activation elu, batch norm,  L2 reg $\alpha$=0.0001\\ \midrule
MaxPooling      & 2x2              & 0.4 dropout                                         \\ \midrule
Fully Connected & 10               & activation softmax                                  \\ \bottomrule
\end{tabular}
\end{adjustbox}
\caption[Neural network architecture exploited in the thesis]{
Model used for experiments related to \textit{CIFAR10} dataset.
Loss function was\textit{ categorical crossentropy}.
}
\label{tab:mymodel}
\end{table}

\section{Network partitioning} 
\label{chap:supermodeling-mlp}

\subsection{Last layers training}

\begin{figure}[H]
\centering
\includegraphics[scale=0.22]{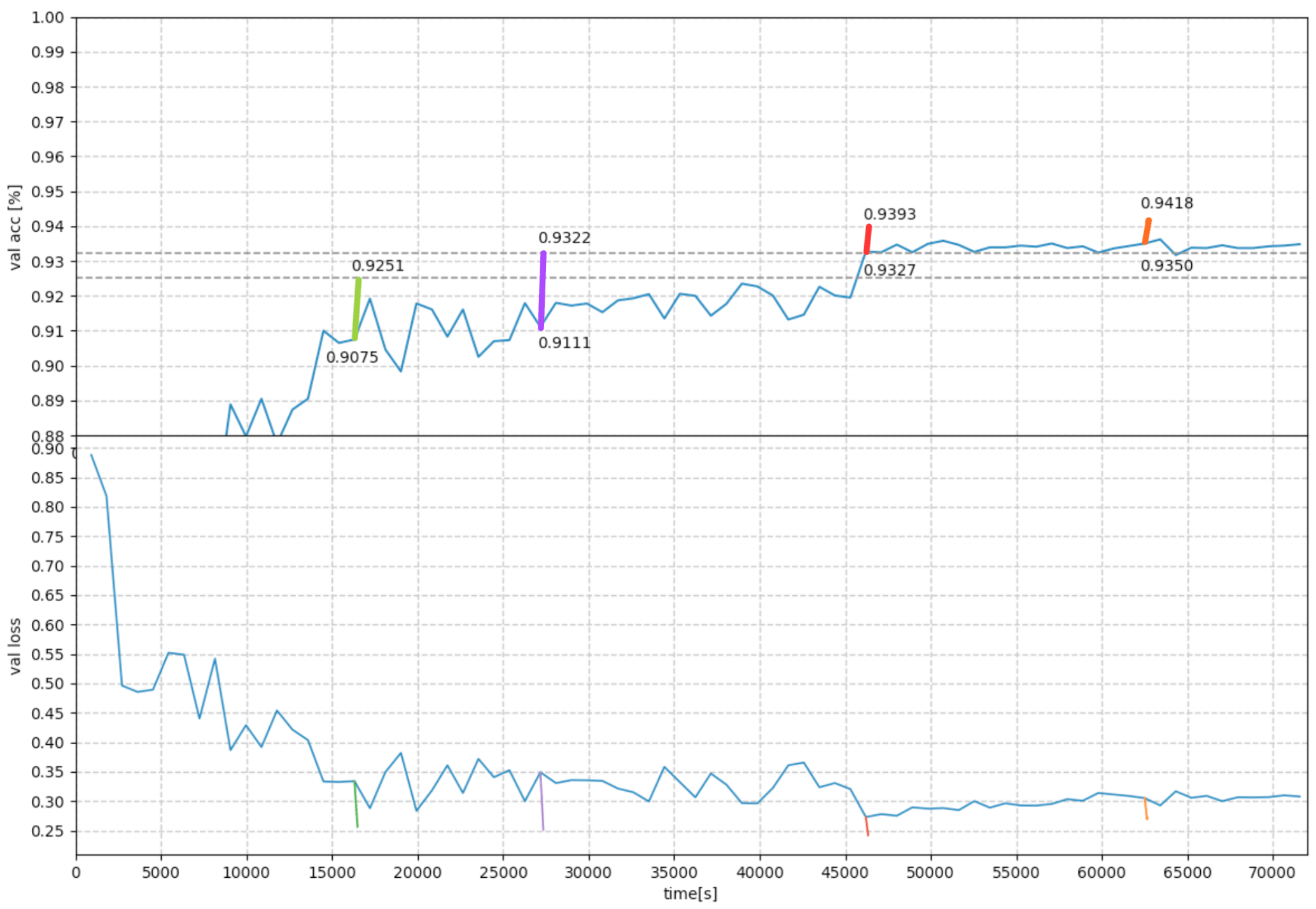}
  \caption[\textit{DenseNet} last layer training]{Validation accuracy and loss for 20 hours(around 80 epochs) of training \textit{DenseNet} model on \textit{CIFAR10} dataset. 
  We applied additional training of only the last layer for snapshots that we captured at 18, 30, 50, and 70 epoch of full network training. The second gain(violet line) replicates the score that was achieved by full network training more than 5h later. The accuracy achieved by the additional boost at 12th hour of training(red line) did not improve within the whole 20 hours of training. Nevertheless, most likely, 20 hours was not enough time for the network to fully converge to its maximum score. The axis is the CPU time spent on computations.} 
\label{fig:densenet}
\end{figure}

Reservoir computing\cite{reservoir_comp} is a learning paradigm that introduces the concept of dividing the system into input, reservoir, and readout components. The main characteristic of such architecture is that only the readout weights are trained, which significantly reduces the computational time. We ported that simple idea into neural networks. We propose the training method, in which most of the time we spend on fitting the whole model and then, we disable all the layers except the last one. The remaining computational budget we use for training just the last, active layer. The intuition behind that approach is the assumption that the first levels of network are responsible for recognition rather simple features, like angles or primitive shapes, hence are not learning significantly in the later stages of the training. 
The preliminary results for such an approach started to look indeed promising, as could be seen in Fig. \ref{fig:densenet}. It illustrates validation accuracy and loss for the training of DenseNet\cite{densenet} architecture on \textit{CIFAR10} dataset and presents the moments when we took snapshots of the network, and we trained only the last layer.

Intrigued by that observation, we started applying additional fitting for the last fully-connected layer, and up to the last two convolutional layers. Each time, we fitted only one level of neurons, when we finished the training of the consequent layers. The motivation behind the descending order was the assumption that the gradients are better when the weights of successor neurons are more convergence to the loss function's optimum. 

We have applied that method for six different pre-trained models from Table \ref{tab:mymodel}. We present the results in Table \ref{tab:supermodeling}. The important note is that when we train the layers in descending order, starting from the last one, we are achieving higher accuracy. In that case, an additional 13 epochs improve the score of each network by 2-5\%.
Having in mind that there is only one layer trained at the time, it tends to overfit in a short amount of time. For that reason, only a few epochs should be run at each stage.

\begin{figure}[H]
\centering
\includegraphics[scale=0.2]{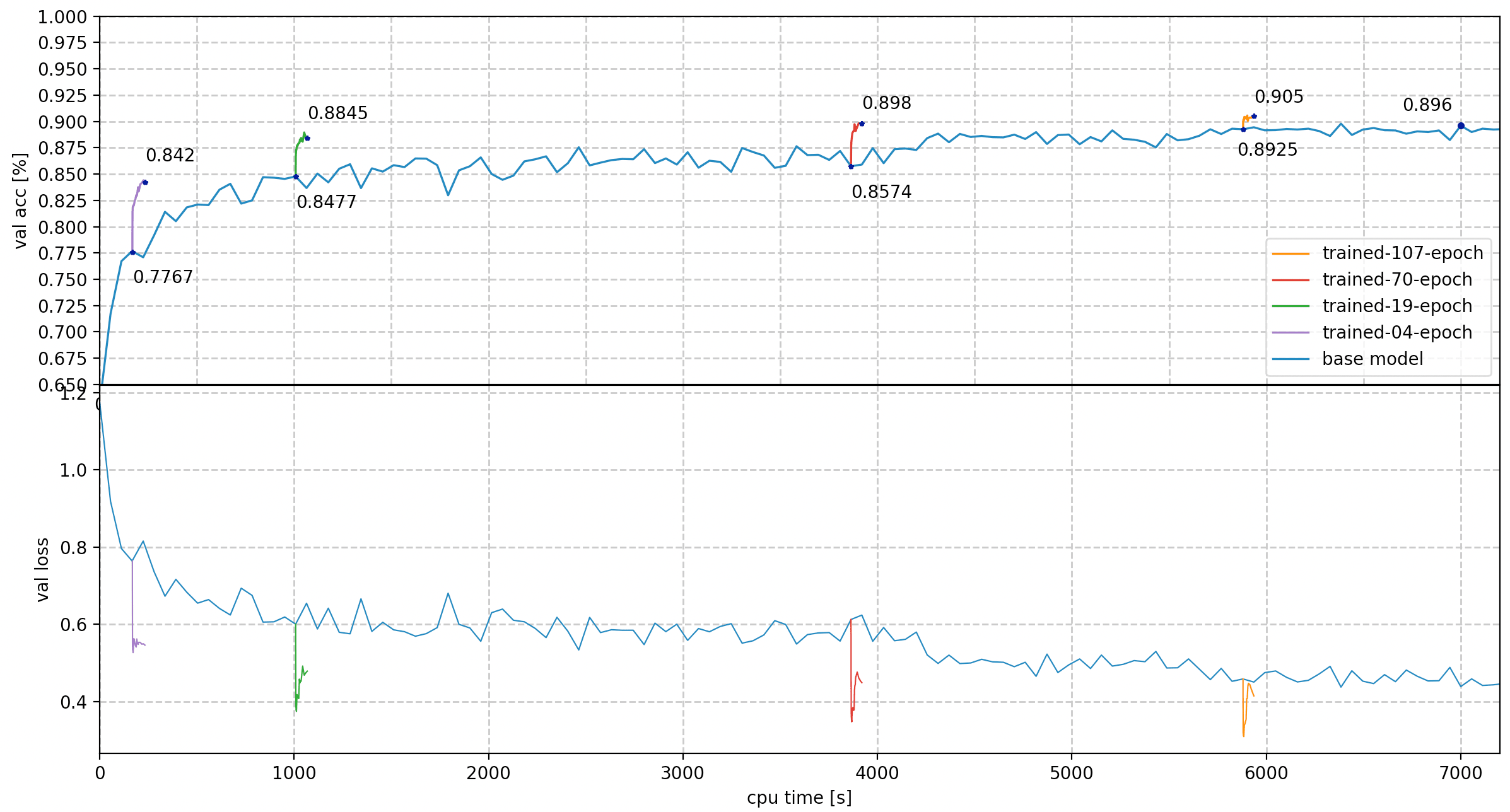}
  \caption[Last layers training for \textit{CIFAR10} dataset]{We have trained the model from Table \ref{tab:mymodel} on \textit{CIFAR10} dataset and saved snapshots of 4,19,70 and 107 epochs. Then, we additionally trained the last four layers (one layer at a time, starting from the last) of the snapshots. We trained each layer for four epochs. The last blue dot indicates the maximum score of the base network without additional last layer training. }
\label{fig:cifar10}
\end{figure}

\begin{figure}[H]
\centering
\includegraphics[scale=0.2]{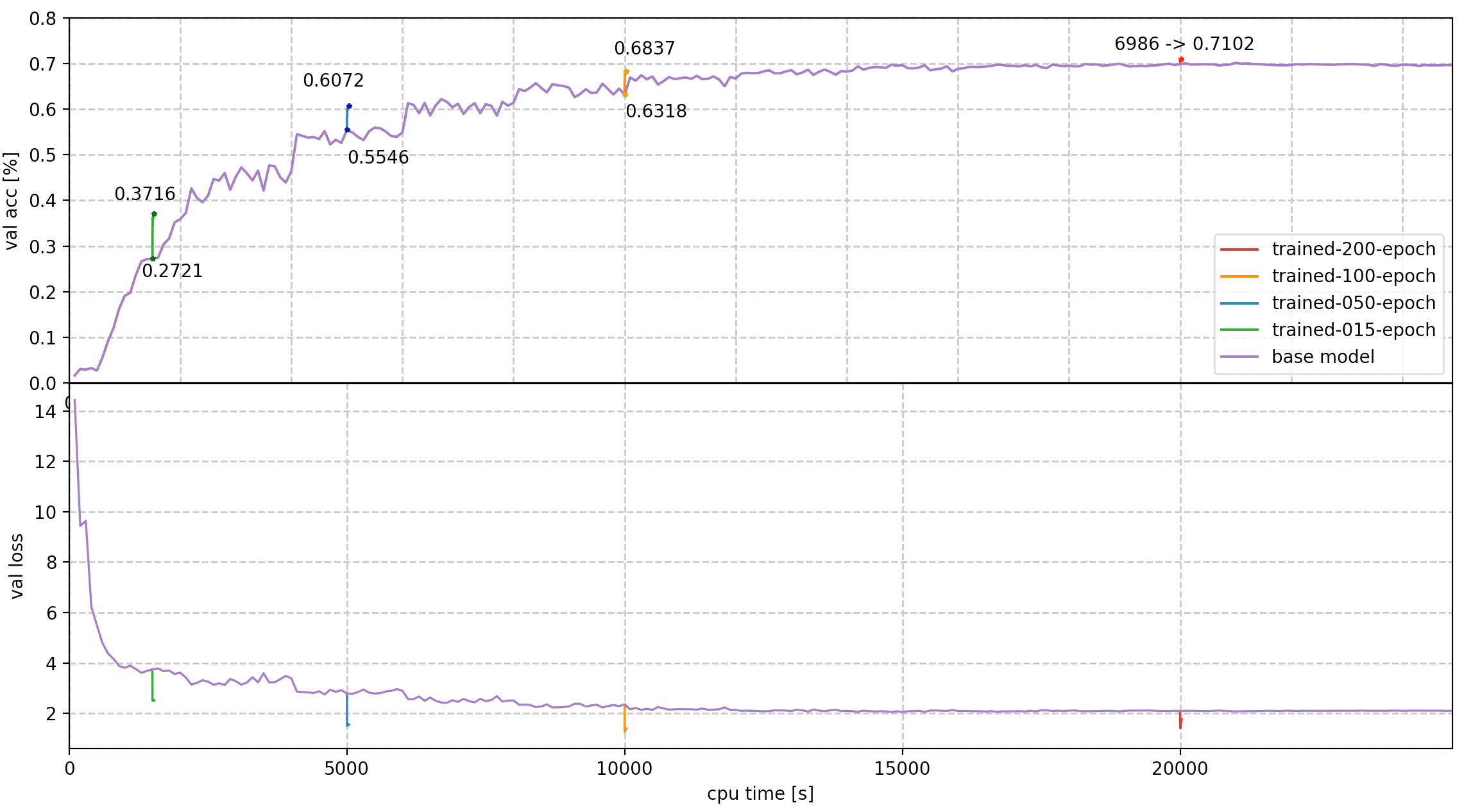}
  \caption[Last layers training for \textit{CIFAR100} dataset]{We trained a \textit{VGG-16}\cite{vgg} model on \textit{CIFAR100} dataset for 250 epochs. We took snapshots in the 15th,50th,100th and 200th epoch. Then, for those snapshots we trained last four layers for 4 epochs each, one layer by one. The last fit improves accuracy for more than 1\%. Very close to SOTA obtained for \textit{VGG-16} (71.56\%) in much shorter time\cite{vgg_sota}.}
\label{fig:cifar100}
\end{figure}

To have a measure of the speedup of that approach, we have compared this method to classical training in Figures \ref{fig:cifar10} and \ref{fig:cifar100} for datasets CIFAR10 and CIFAR100 respectively.  As we can see, the additional accuracy gain is slowly decreasing over the training time.

\begin{figure}[H]
\centering
\includegraphics[scale=0.3]{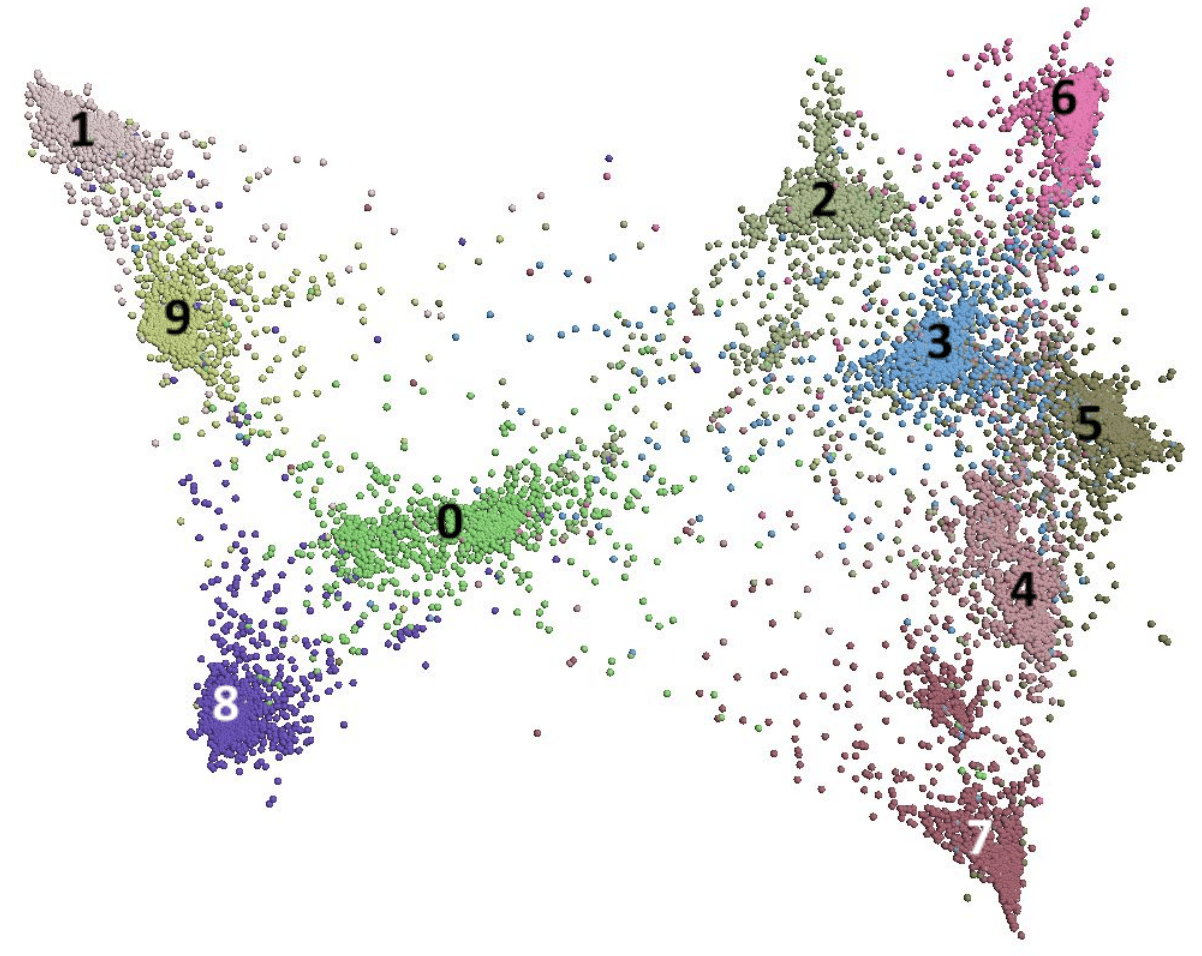}
  \caption[\textit{DenseNet} predictions from penultimate layer]{We trained \textit{DenseNet} model to achieve 93.62\% on \textit{CIFAR10} dataset. Then, we visualized\cite{embedding} the outputs of the testing set from penultimate layer; each example was 980 dimensional vector. The numbers indicate \textit{CIFAR10}' classes.} 
\label{fig:densenet_classes}
\end{figure}

\begin{figure}
\begin{subfigure}{.5\textwidth}
  \centering
  \includegraphics[width=1.0\linewidth]{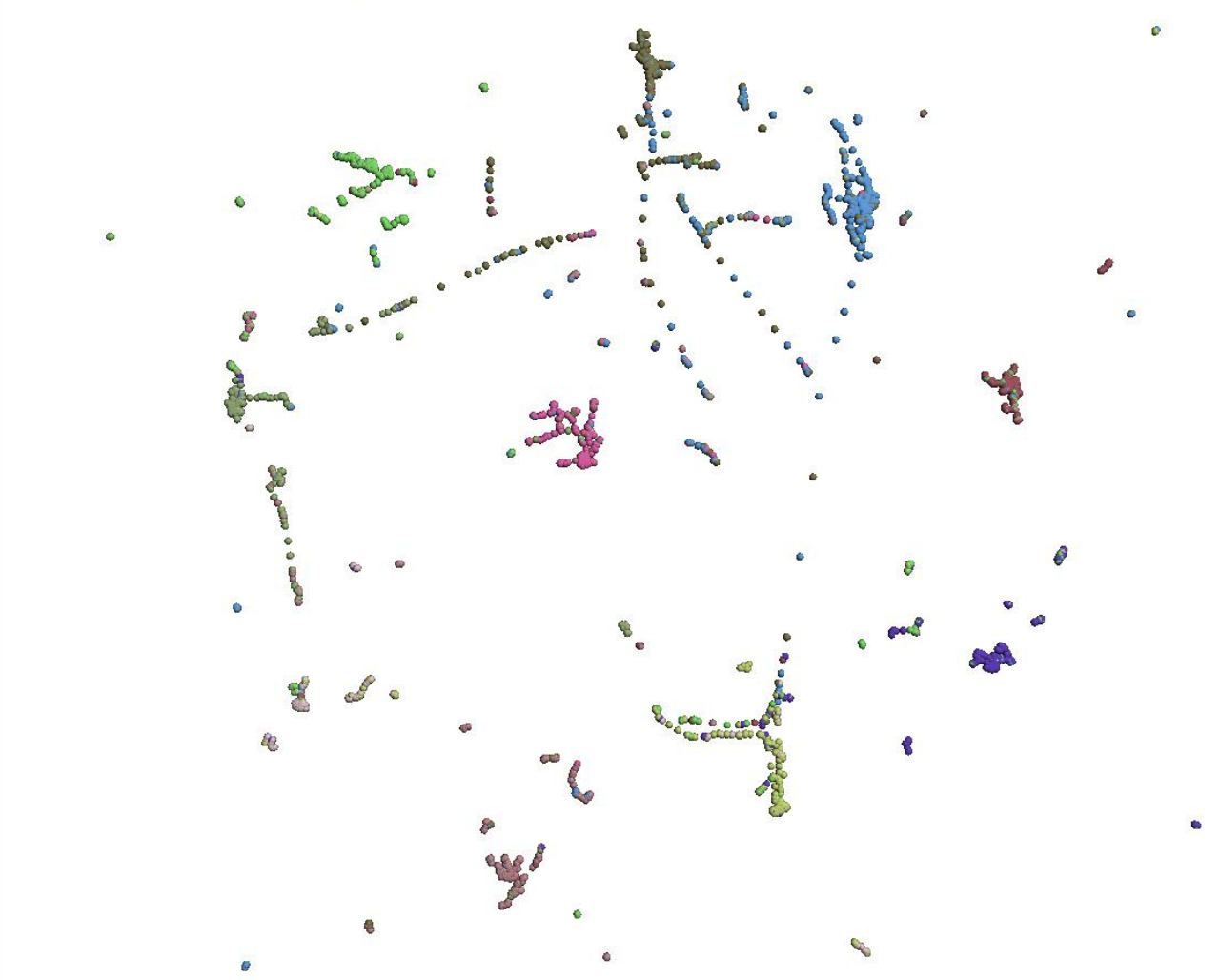}
  \caption{before last layer re-train.}
  \label{fig:sfig1}
\end{subfigure}
\begin{subfigure}{.5\textwidth}
  \centering
  \includegraphics[width=1.0\linewidth]{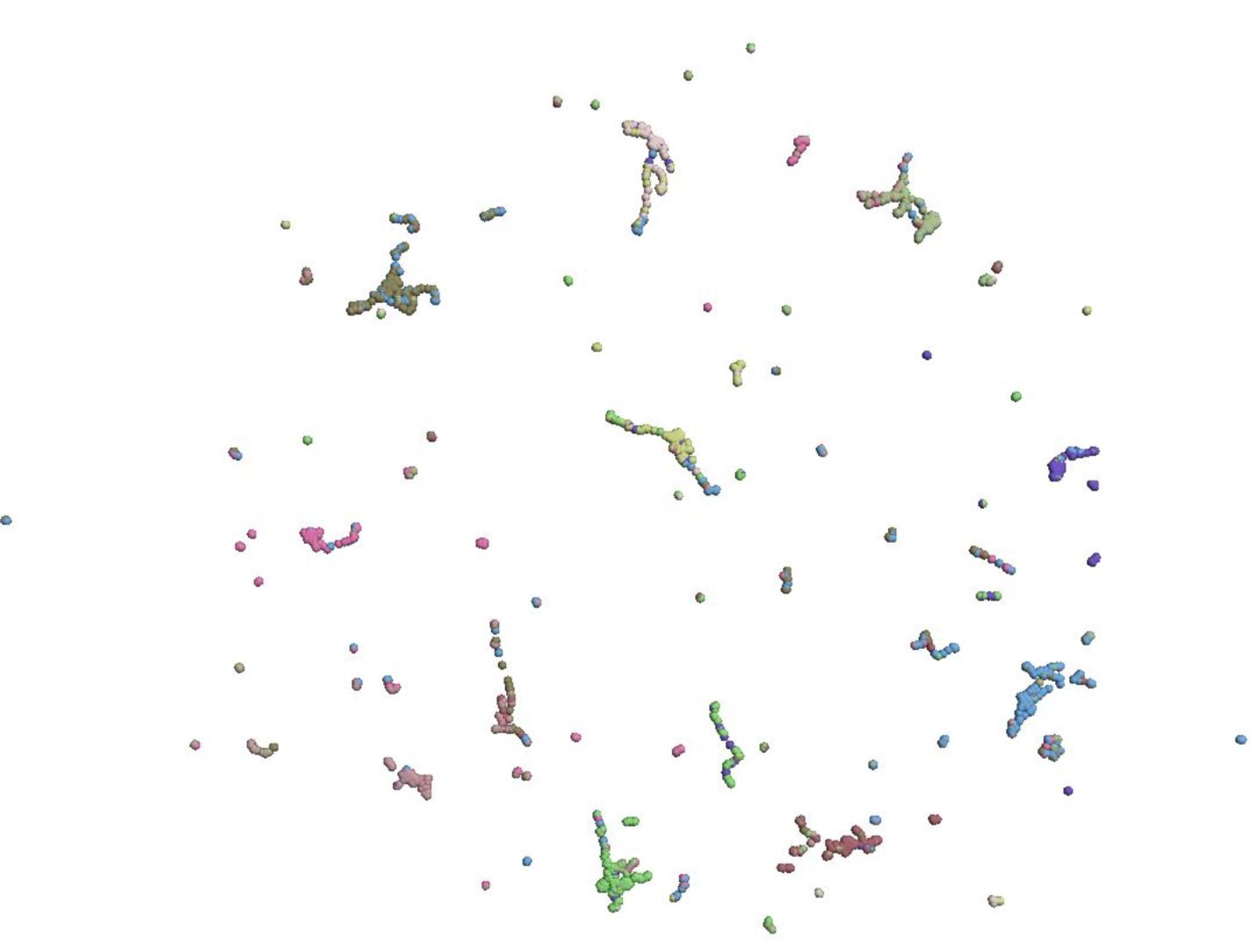}
  \caption{after last-layer retrain. }
  \label{fig:sfig2}
\end{subfigure}
\caption[\textit{DenseNet} predictions with last layer re-trained]{
This Figure presents the 2D embedding of the model's (from Figure \ref{fig:densenet_classes}) predictions for the test set before and after re-training of the last layer. The 5-nn metrics\cite{embedding} are 0.906 and 0.912, respectively.

}
\label{fig:densenet_last_classes}
\end{figure}

For a better understanding of the characteristics of the last layer's training, we were using visualization methods\cite{embedding} of the network's predictions.
Firstly, in Fig. \ref{fig:densenet_classes}, we passed the whole testing set through a pre-trained DenseNet network, and we captured the outputs from the penultimate layer. We observe that this fraction of the model is enough to distinguish the classes correctly. In Fig. \ref{fig:densenet_last_classes}, we illustrate predictions of the whole network before, and after last layer re-train. After the train, the clusters seem to be more concentrated, which confirms the improving properties of this method. Those results also present the considerable difference between the outputs of the last and penultimate layer, which may be a sign of the significant role in the classification of the final level of neurons.

To conclude, "Freezing" all layers except the last ones speeds up the training significantly. It can be used as a quick accuracy boost when we have limited resources. It is a fascinating observation, and it is worth to do a separate study around that method.

\begin{table}[H]
\centering
\begin{adjustbox}{width=1.1 \textwidth}
\begin{tabular}{|l|l|l|l|l|l|}
\hline
        & base accuracy & \begin{tabular}[c]{@{}l@{}}last layer \\ (3 epochs)\end{tabular} & \begin{tabular}[c]{@{}l@{}}2th layer from the end\\ (3 epochs)\end{tabular} & \begin{tabular}[c]{@{}l@{}}3th layer from the end\\ (3 epochs)\end{tabular} & \begin{tabular}[c]{@{}l@{}}4th layer from the end\\ (4 epochs)\end{tabular} \\ \hline
model 1 & 0.8548        & 0.8843                                                           & 0.8929                                                                  & 0.8999                                                                  & 0.9035                                                                  \\ \hline
model 2 & 0.8434        & 0.8791                                                           & 0.8892                                                                  & 0.8944                                                                  & 0.8975                                                                  \\ \hline
model 3 & 0.8619        & 0.8789                                                           & 0.8854                                                                  & 0.8890                                                                  & 0.8949                                                                  \\ \hline
model 4 & 0.8489        & 0.8812                                                           & 0.8915                                                                  & 0.9007                                                                  & 0.9012                                                                  \\ \hline
model 5 & 0.8636        & 0.8824                                                           & 0.8913                                                                  & 0.8943                                                                  & 0.8978                                                                  \\ \hline
model 6 & 0.8739        & 0.8871                                                           & 0.8939                                                                  & 0.8927                                                                  & 0.8965                                                                  \\ \hline
\end{tabular}
\end{adjustbox}
\caption[Training layers one-by-one]{
Validation accuracy for one-by-one layer training of six different models (\ref{tab:mymodel}) trained on \textit{CIFAR10}. We can see that accuracy improves by 2-5\% in just 13 additional epochs.
}
\label{tab:supermodeling}
\end{table}

\subsection{Partitioning}

\begin{figure}[H]
\centering
\includegraphics[scale=0.7]{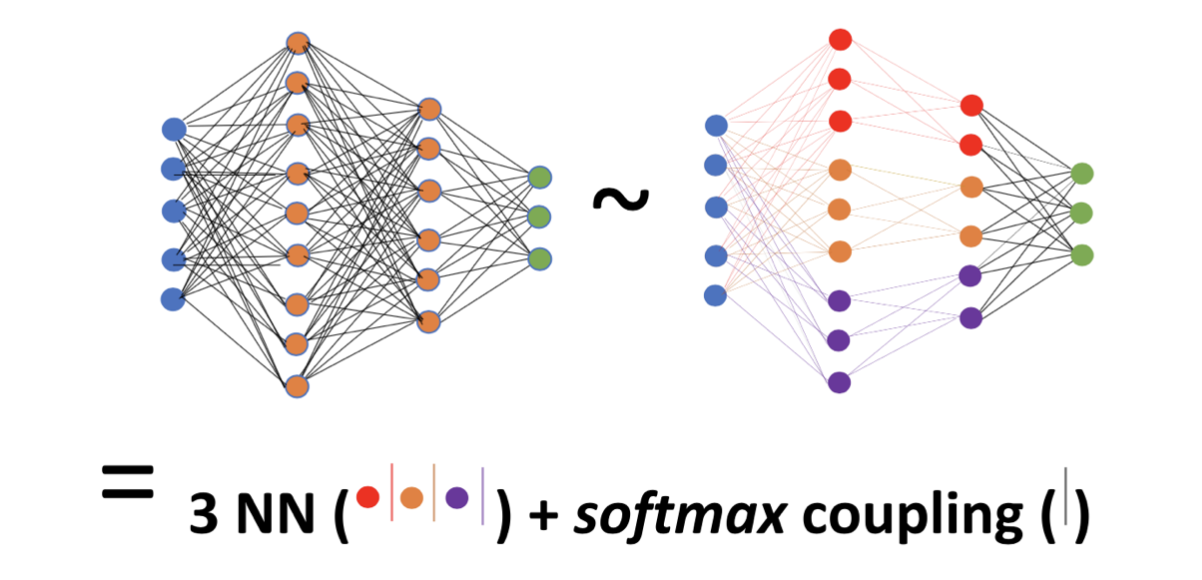}
  \caption[Network partitioning]{If we remove some connections in a MLP network, it could be tread as an ensemble.}
\label{fig:coupling}
\end{figure}

\begin{figure}[H]
\centering
\includegraphics[scale=0.5]{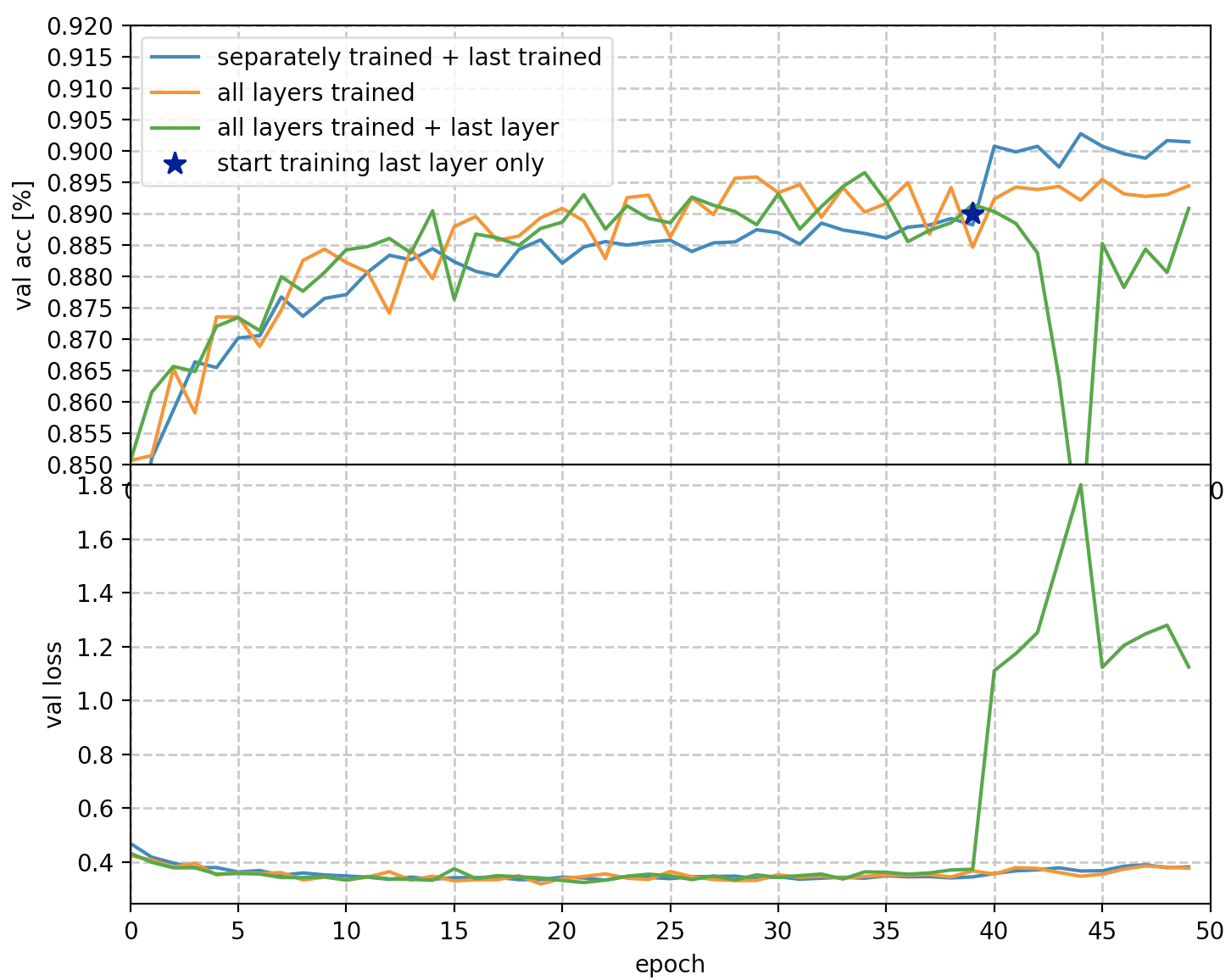}
  \caption[Different methods for last layer training for an ensemble]{
  Three different approaches for training architecture proposed in the Fig. \ref{fig:coupling} trained on \textit{Fashion MNIST} dataset.
  The network consisted of four layers: (240, 560, 560,10) neurons, and we divided it into four subbranches. We used \textit{dropout} with probability rate set to 0,1 between layers, \textit{ReLu} as activation function and L2 regularization with $\alpha$=0.001.}
 \begin{enumerate} [I.]
\item orange line - we trained the whole network for 50 epochs.
\item green line - we trained the whole network for 39 epochs and then ten epochs for only the last layer.
\item blue line - we trained four independent networks for 39 epochs, and then we merged them with the last layer, which we additionally fit for ten epochs.
\end{enumerate}
\label{fig:supermodels_compare}
\end{figure}

\begin{figure}[H]
\centering
\includegraphics[scale=0.8]{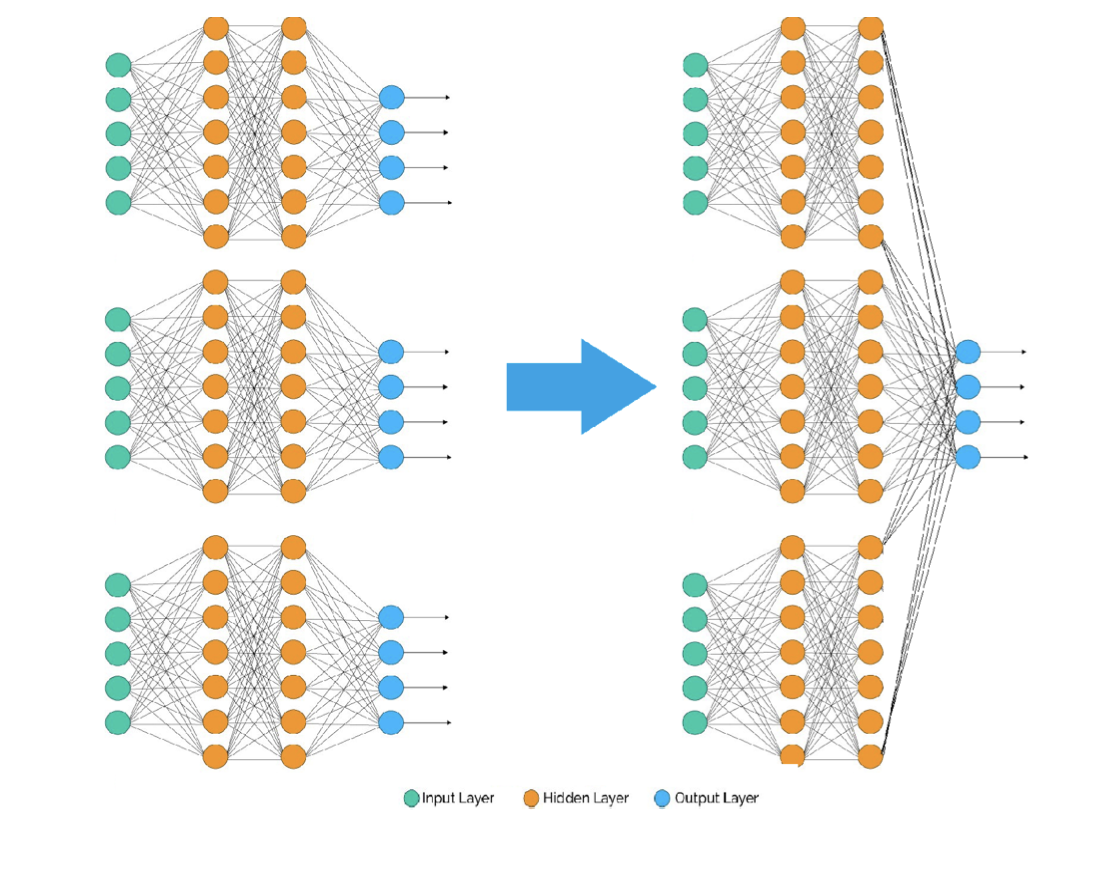}
  \caption[Ensembling with sub-models coupling]{Construction of \textit{SuperModel} architecture from 3 sub networks. After the merge, we additionally re-trained the last layer for short time with training data.}
\label{fig:supernet}
\end{figure}

Let us consider a dense network on which some internal connections between neurons had been removed. Therefore, we can treat such a model as an ensemble of multiple smaller networks.
Fig. \ref{fig:coupling} should be helpful in understanding that observation. We have tested three different approaches for training the NN architecture shown in Fig. \ref{fig:supermodels_compare}, by training:
\begin{enumerate} [I.]
\item the whole network,
\item the whole network and additionally the last layer,
\item the subbranches independently and then merged them with the last layer, which is additionally trained.
\end{enumerate}
Whereas the first two methods seem to yield similar or slightly better results than training a single network, \textbf{the SuperModel with re-trained last layer improves the performance (Fig. \ref{fig:supermodels_compare})}. It may lead to a more generalized idea of creating efficient networks.
The construction of the architecture is as follows: multiple dense networks that we trained on the same data are merged with the last, softmax layer. For that purpose, we are copying the corresponding weights from the last layers of subnetworks to the merged model and we are initializing biases to the average values of corresponding biases. Fig. \ref{fig:supernet} shows how sample supermodel is constructed.
Then, inspired by the results from the previous chapter, we propose to additionally re-train the last layer with \textit{ the same} training data for few(up to 10) epochs. From now on, each \textit{SuperModel} in my thesis refers to the ensemble with the re-trained softmax layer.
In my thesis we use \textit{ReLu} for internal activations and \textit{softmax} as output function for fully-connected networks.
 The reason behind limiting the architecture type to multilayer perceptrons(\textit{MLP}) in this chapter is that it is possible to "cut alongside" such models and therefore combine the components in \textit{SuperModel} structures. The final model has the same amount of neurons in each layer but less connections, which means less trainable parameters. That approach gives many opportunities to compare the ensembles with the classical networks of the same size. 

As a next point of our research, we have divided \textit{Fashion MNIST} dataset into training, validation, and testing set and created three different sizes of dense model. For each size, we have divided the MLP into 2, 3, 4, and 6 pre-trained subnetworks and created \textit{SuperModel}. We finished the training of each submodel by stopping it when the validation loss reached a specified threshed. The Table \ref{tab:cutmodel} presents the accuracy and loss for the testing dataset with measured CPU time. Those preliminary results demonstrate that the \textit{SuperModel} indeed achieves higher testing accuracy for bigger sizes of the architecture than corresponding single models. Such an observation seems quite intuitive; ensembles result in better performance but consume more CPU resources for training. In the rest of this chapter, we would like to analyze the ensemble's properties more deeply and focus on how many benefits \textit{SuperModel} can give.
\begin{table}[]
\centering
\begin{adjustbox}{width=0.8 \textwidth}
\begin{tabular}{|l|l|l|l|l|l|}
\hline
\multicolumn{6}{|c|}{\textit{Small network}}                                                          \\ \hline
\textit{subnetworks} & \textit{1}        & \textit{2} & \textit{3}        & \textit{4} & \textit{6}        \\ \hline
\textit{test acc}    & \textbf{0.8958}   & 0.8906     & 0.8962            & 0.8943     & 0.8893            \\ \hline
\textit{test loss}   & \textbf{0.3194}   & 0.344      & 0.3451            & 0.333      & 0.346             \\ \hline
\textit{cpu time}    & \textbf{00:01:49} & 00:02:04   & 00:03:19          & 00:03:48   & 00:06:01          \\ \hline
\multicolumn{6}{|c|}{\textit{Medium network}}                                                         \\ \hline
\textit{subnetworks} & \textit{1}        & \textit{2} & \textit{3}        & \textit{4} & \textit{6}        \\ \hline
\textit{test acc}    & 0.8942            & 0.8914     & \textbf{0.8971}   & 0.8927     & 0.895             \\ \hline
\textit{test loss}   & 0.349             & 0.3445     & \textbf{0.3506}   & 0.370      & 0.3659            \\ \hline
\textit{cpu time}    & 00:03:53          & 00:01:51   & \textbf{00:02:52} & 00:03:34   & 00:06:24          \\ \hline
\multicolumn{6}{|c|}{\textit{Large network}}                                                          \\ \hline
\textit{subnetworks} & \textit{1}        & \textit{2} & \textit{3}        & \textit{4} & \textit{6}        \\ \hline
\textit{test acc}    & 0.891             & 0.8983     & 0.8973            & 0.8953     & \textbf{0.8994}   \\ \hline
\textit{test loss}   & 0.315             & 0.4026     & 0.404             & 0.405      & \textbf{0.424}    \\ \hline
\textit{cpu time}    & 00:06:50          & 00:04:57   & 00:05:37          & 00:06:29   & \textbf{00:08:59} \\ \hline
\end{tabular}
\end{adjustbox}
\caption[Network partitioning]{
Test accuracy and loss for networks that we partitioned into 2,3,4, and 6 subnetworks. Each root model consisted of four layers:
}
\begin{enumerate} [a)]
\item Small network: 360, 840, 840 and 10 neurons. We set \textit{dropout} between layers to 0.3 for single network and 0.2 for subnetworks. We finished the training of single network when loss stopped to decrease for consecutive 10 epochs, and for 5 epochs for subnetworks.
\item Medium network: 720, 1680, 1680 and 10 neurons. We set \textit{dropout} between layers to 0.3 for single network and 0.2 for subnetworks. We finished the training of single network when loss stopped to decrease for consecutive 10 epochs, and for 5 epochs for subnetworks.
\item Large network: 1200, 2800, 2800 and 10 neurons. We set \textit{dropout} between layers to 0.7 for single network and 0.3 for subnetworks. We finished the training of single network when loss stopped to decrease for consecutive 20 epochs, and for 10 epochs for subnetworks.
\end{enumerate}

\label{tab:cutmodel}
\end{table}

\subsection{Loss function dilemma}
\begin{figure}[H]
\centering
\includegraphics[scale=0.6]{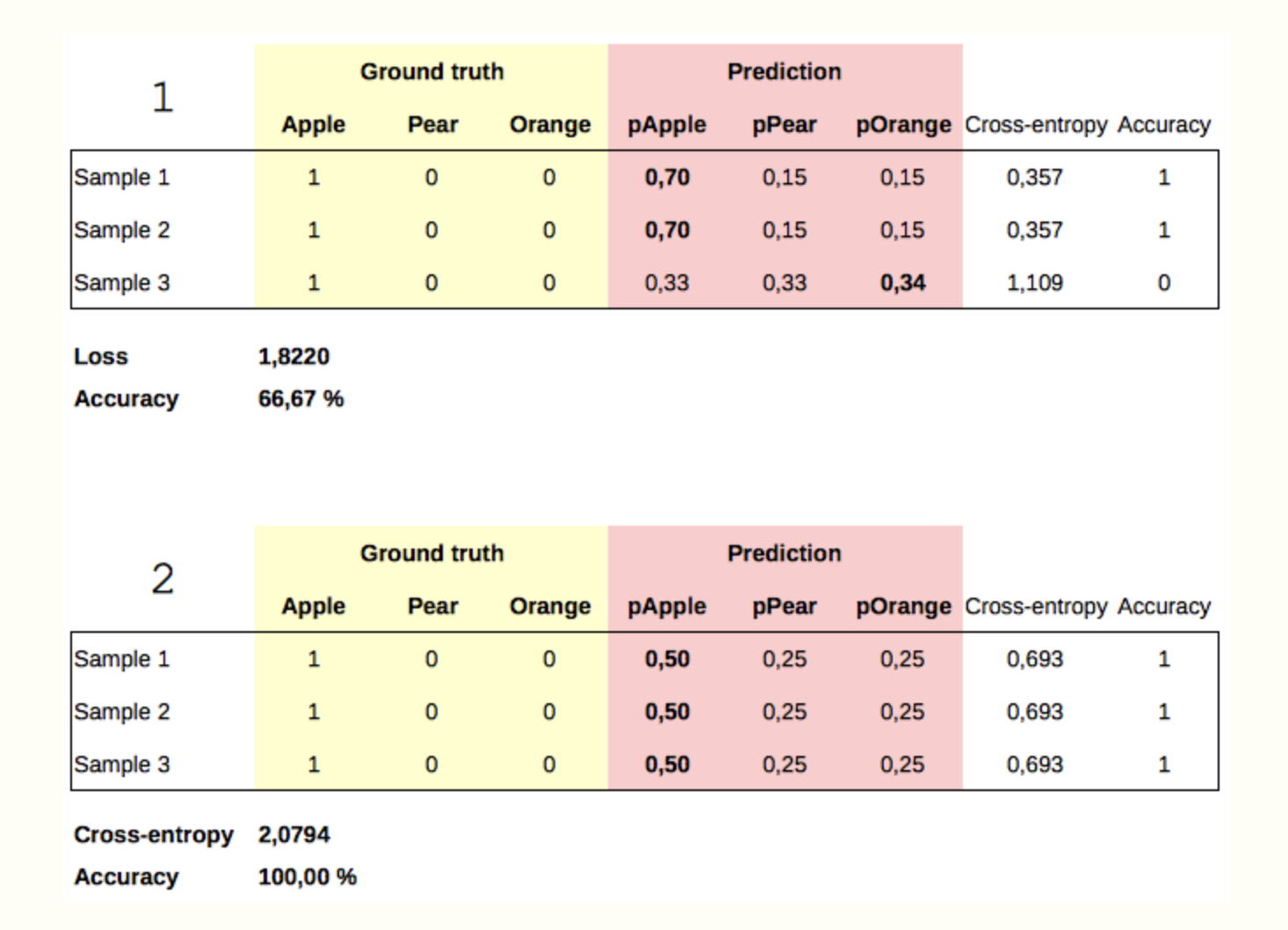}
  \caption[Loss-accuracy relation]{Example that presents the scenario that the lower loss of \textit{softmax} function does not always mean the lower score in general. Even though the second predictor outputs lower probabilities (it's  "less sure"), his final accuracy is higher\cite{loss-example}.}
\label{fig:exampleloss}
\end{figure}

\begin{figure}[H]
\centering
\includegraphics[scale=0.5]{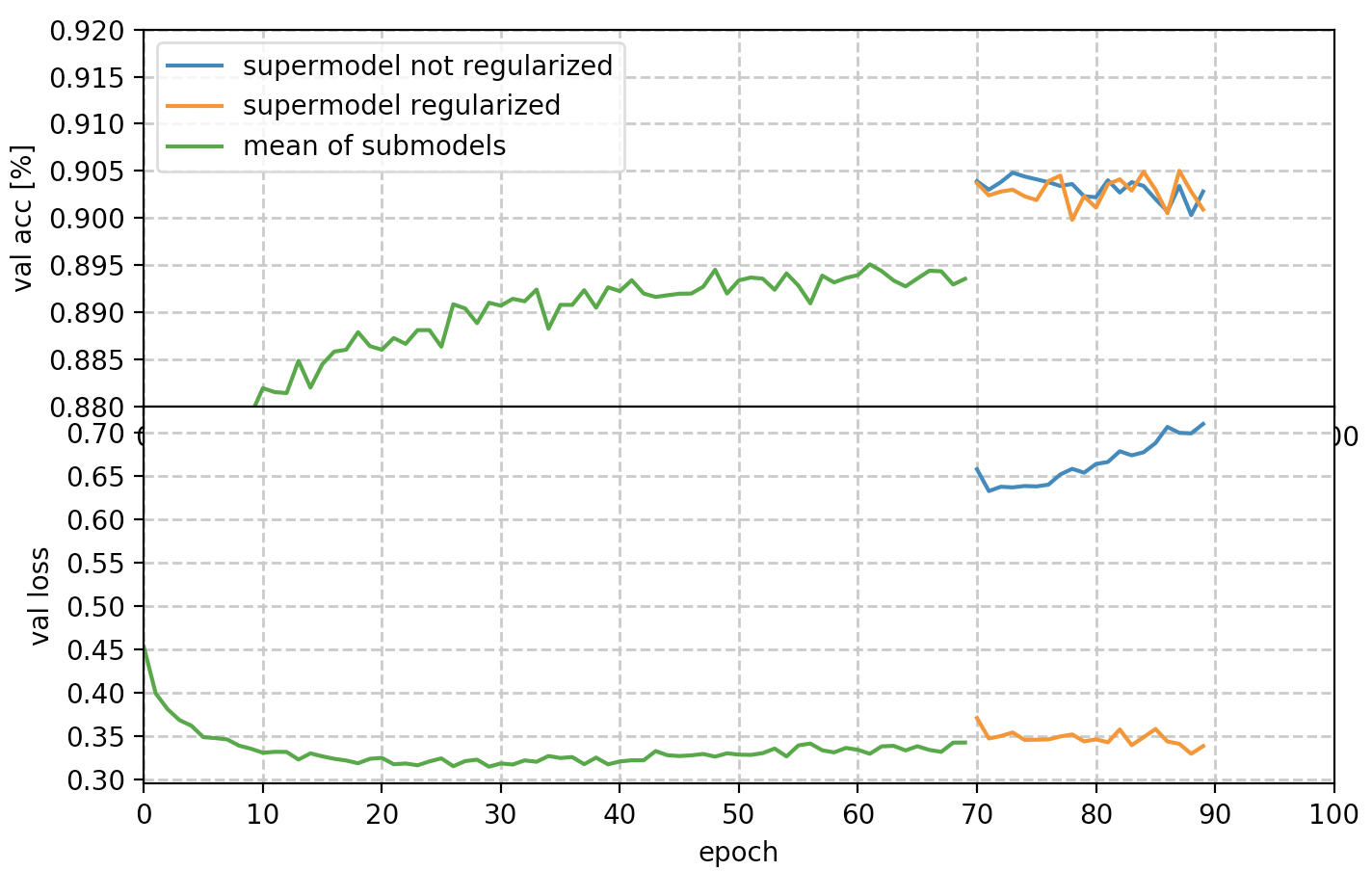}
  \caption[Regularized ensemble]{\textit{SuperModel} created from models trained on \textit{Fashion MNIST} Dataset. The submodels were fully-connected networks with four layers: (120, 280, 280, 10) neurons. For both trainings, we copied the corresponding weights  directly from submodels into \textit{Supermodel}. L2 regularization added to weights and bias during \textit{SuperModel's} training reduces the error without accuracy drop(orange line). 
  }
\label{fig:highloss}
\end{figure}

During our experiments around the \textit{Supermodeling} on different datasets, we have always observed one striking issue.
Although the validation accuracy of such an ensemble is higher than the corresponding submodels, the testing/validation loss tends to be higher as well (Table \ref{tab:cutmodel}). The overfitting is the most common answer to that behavior. We have a slightly different interpretation for that phenomenon. In our opinion, it results from the loosely coupled nature between \textit{softmax} loss function and its finally predicted class. Figure \ref{fig:exampleloss} may help to understand this relation. There may be two scenarios happening at the same time. Firstly, the ensemble is overconfident of its predictions which results in increased error loss for incorrect guesses. At the same time, the border examples (the examples that networks is not sure about) are predicted better, which improves the score. If the first case happens often enough, the loss function will grow with untouched accuracy, which is additionally "bumped" by the second scenario. In order to prove this hypothesis, we measured the basic properties of \textit{SuperModel} and its corresponding subnetworks in Table \ref{tab:losscompare}. Indeed the \textit{SuperModel} has a higher mean loss, but 90\% of the examples reach a minimal error. When we interpret this together with the bigger standard deviation, we can conclude that the mean error is increased mostly by the remaining 10\% of guesses. Overconfidence means that the output probabilities are closer to the edge values (0 and 1); therefore, the incorrect predictions lead to very high cross-entropy. Moreover, the 95th percentile shows much higher value which we can explain by the fact that it is more than the accuracy of the \textit{SuperNet}(91\%) thus the remaining ~4\% must be wrong guesses that generate very high error (Table \ref{tab:losscompare}).
It is known that overconfident predictions may be the symptom of overfitting\cite{reg_entropy}. However, we need to keep in mind that creating an ensemble we are actually building a new model with different capabilities, whereas overfitting is related to the \textit{process} of training a single model. In our experiments, the regularization method of L2 penalty added to weights in the stage of training the last layer reduces the loss significantly. On the other hand, it usually had smaller final accuracy of the ensemble in comparison to non-regularized ensembles.
To have a better insight, Table \ref{tab:rescale} presents validation accuracy and loss for training \textit{SuperModel's} \textit{softmax}'s layer on two different datasets. In Table \ref{tab:rescale}, we compared three different initializations of last layer's weights - \textit{a)} when the coefficients are initialized in random, \textit{b)} when we directly copied the corresponding weights from the submodels and \textit{c)} when we copied the weights and additionally re-scaled them by dividing by an arbitrary factor. The intuition behind the last case comes from the conclusion that the overconfidence, which we observed, is occurring when the weights reach high values. Building \textit{SuperModel}, we are combining connections for the last layer from multiple submodels. If the activations for the same feature are similar in each model (e.g., each corresponding neurons give similarly negative value for a particular example) then the final value is proportionally higher. To prevent this effect, we have reduced the coefficients with division operation. We have repeated the experiment on models of different sizes and we very often came up with the same conclusion that is the highest score (accuracy and loss) achieves the non-regularized training of the last layer on which the coefficients were copied and down-scaled. However, we did not come up with a standardized way of normalizing the weights, and we were choosing the factors experimentally. The less intuitive is the fact that the best division factor was sometimes much bigger than the number of subnetworks(e.g., 60), depending on the model.
Nevertheless, as shwon in Fig. \ref{fig:highloss}, the difference is not that huge, and in some cases, choosing regularization parameters carefully still decreased the error, without impacting the precision of the outputs. 

In our opinion, using \textit{Supermodeling} with randomly initialized weights and regularization may be a feasible way of increasing the network's performance enough, so that we can keep proper accuracy/loss balance. We think that coping the weights from submodels should not be a necessary step and we suspect a trivial reason why it worked in our experiments. Each time, we were initializing a new optimizer for the last layer training, which forced the coefficients to move dramatically, hence their initial state was not that important. 
The problem of the \textit{Supermodel}'s high confidence and its response to regularization is tightly related to the level of variance of the submodels that we combine. This is the topic that we try to address in the next paragraph.

\begin{table}[H]
\centering
\begin{tabular}{|l|l|l|l|l|l|}
\hline
                      & accuracy & loss mean & loss std    & loss 90th        & loss 95th \\ \hline
6 Submodels (average) & 0.89315  & 0.65722   & 2.4633 & \textbf{0.92668} & 4.4884    \\ \hline
SuperNet              & 0.9103   & 1.0976    & 3.8482 & \textbf{0.06541} & 16.11809  \\ \hline
\end{tabular}
\caption[\textit{Supermodeling} and loss measurements]{
Evaluation of six models and their \textit{SuperModel} on the validation set.
Each submodel was a fully-connected network that we trained on \textit{FMNIST} dataset and had the architecture of four layers(200, 466, 466 and 10 neurons).
\textit{SuperModel} achieves better accuracy despite the higher mean loss.
However, the 90\% of the predictions have small errors. Based on this fact, one can conclude that the high mean value results from very few predictions that were overconfident and incorrect.}
\label{tab:losscompare}
\end{table}

\begin{table}[H]
\centering
\begin{adjustbox}{width=1.0 \textwidth}
\begin{tabular}{|l|l|l|l|
>{\columncolor[HTML]{FFFFFF}}l |
>{\columncolor[HTML]{FFFFFF}}l |
>{\columncolor[HTML]{FFFFFF}}l |}
\hline
                                                                  & \multicolumn{3}{l|}{\cellcolor[HTML]{EFEFEF}Fashion MNIST}                                                                                                                                              & \multicolumn{3}{l|}{\cellcolor[HTML]{EFEFEF}TNG}                                                                                                                                                                  \\ \cline{2-7} 
                                                                  &                                                                          & \multicolumn{2}{l|}{copied weights}                                                                                          & \cellcolor[HTML]{FFFFFF}                                                                         & \multicolumn{2}{l|}{\cellcolor[HTML]{FFFFFF}copied weights}                                                                  \\ \cline{3-4} \cline{6-7} 
\multirow{-3}{*}{}                                                & \multirow{-2}{*}{\begin{tabular}[c]{@{}l@{}}random \\ init\end{tabular}} & \begin{tabular}[c]{@{}l@{}}downscaled\\\end{tabular}          & not scaled                                               & \multirow{-2}{*}{\cellcolor[HTML]{FFFFFF}\begin{tabular}[c]{@{}l@{}}random \\ init\end{tabular}} & \begin{tabular}[c]{@{}l@{}}downscaled\\ \end{tabular}         & not scaled                                               \\ \hline
non-regularized                                                   & \begin{tabular}[c]{@{}l@{}}0.9122/\\ 0.6401\end{tabular}                 & \textbf{\begin{tabular}[c]{@{}l@{}}0.9132/\\ 0.7633\end{tabular}} & \begin{tabular}[c]{@{}l@{}}0.9112/\\ 1.1066\end{tabular} & \begin{tabular}[c]{@{}l@{}}0.8385/\\ 0.9216\end{tabular}                                         & \textbf{\begin{tabular}[c]{@{}l@{}}0.8502/\\ 0.6651\end{tabular}} & \begin{tabular}[c]{@{}l@{}}0.8458/\\ 1.3965\end{tabular} \\ \hline
\begin{tabular}[c]{@{}l@{}}\textit{L2} penalty,\\ alfa=0.01\end{tabular}   & \begin{tabular}[c]{@{}l@{}}0.9105/\\ 0.4187\end{tabular}                 & \begin{tabular}[c]{@{}l@{}}0.9088/\\ 0.4295\end{tabular}          & \begin{tabular}[c]{@{}l@{}}0.9089/\\ 0.5079\end{tabular} & \begin{tabular}[c]{@{}l@{}}0.8424/\\ 0.6988\end{tabular}                                         & \begin{tabular}[c]{@{}l@{}}0.8484/\\ 0.6433\end{tabular}          & \begin{tabular}[c]{@{}l@{}}0.8410/\\ 0.7225\end{tabular} \\ \hline
\begin{tabular}[c]{@{}l@{}}\textit{L2} penalty, \\ alfa=0.001\end{tabular} & \begin{tabular}[c]{@{}l@{}}0.9113/\\ 0.5207\end{tabular}                 & \begin{tabular}[c]{@{}l@{}}0.9102/\\ 0.5599\end{tabular}          & \begin{tabular}[c]{@{}l@{}}0.9105/\\ 0.5705\end{tabular} & \begin{tabular}[c]{@{}l@{}}0.8429/\\ 0.8208\end{tabular}                                         & \begin{tabular}[c]{@{}l@{}}0.8497/\\ 0.6678\end{tabular}          & \begin{tabular}[c]{@{}l@{}}0.8467/\\ 1.4009\end{tabular} \\ \hline
\end{tabular}
\end{adjustbox}
\caption[Last layer initializations of the NN ensemble ]{
We have compared a few different approaches for \textit{SuperModel's} last layer initialization on two datasets.
The presented values are \textit{Validation accuracy}/\textit{Validation error} for the best score achieved(for about 20 epochs of the softmax layer training).
For \textit{Fashion MNIST} six subnetworks were pretrained, each 200, 466, 466 and 10 neurons.
For \textit{TNG} dataset six subnetworks were pre-trained, each 30, 30 and 10 neurons. For the downscale, we chose the factors experimentally. The worst loss achieves the non-regularized weights directly copied from the submodels.}
\label{tab:rescale}
\end{table}



\subsection{Best submodels}

\begin{figure}[H]
\centering
\includegraphics[scale=0.5]{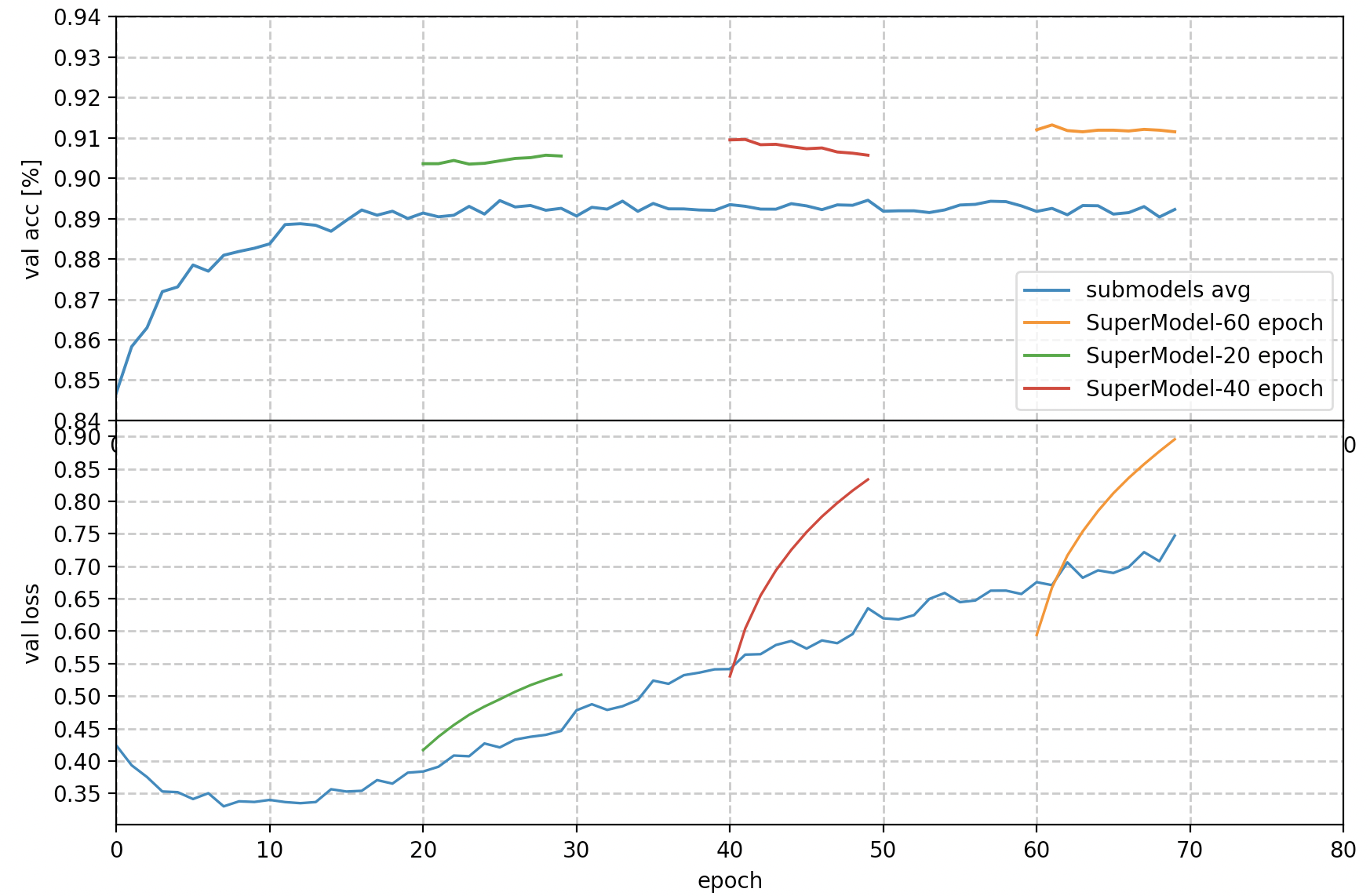}
  \caption[\textit{Supermodel} at different epochs]{\textit{The SuperModel} created at 20, 40 and 60th epoch of training 6 models on \textit{Fashion MNIST} dataset. Each model had architecture of four layers; 40 60, 60 and 10 neurons. Even though it seems that the base models are getting overfit (regularization was not used), the \textit{Supermodel} created at the latest stage of the training appears to have the highest score.}
\label{fig:overconfidence}
\end{figure}

A good entry point to this paragraph will be Fig. \ref{fig:overconfidence}.

For that experiment, we have pre-trained six sub-models on \textit{Fashion MNIST} data.
We were building a \textit{SuperModel} at 20, 40 and 60th epoch of the models' training.
Unusual is the fact that the accuracy achieved by the ensemble from the 60th epoch has the higher score than one from the 20th epoch, even though the submodels have very similar accuracy in both points. Moreover, it seems that the base models are getting overfitted, as the accuracy is stable, but loss slowly increases. Our explanation for this phenomenon is that the ensemble has more accurate guesses when the submodels are more overconfident. We have plotted loss outputs for each validation example, respectively for one model from 20 epoch and from the 60th one.

\begin{figure}
\begin{subfigure}{.5\textwidth}
  \centering
  \includegraphics[width=.9\linewidth]{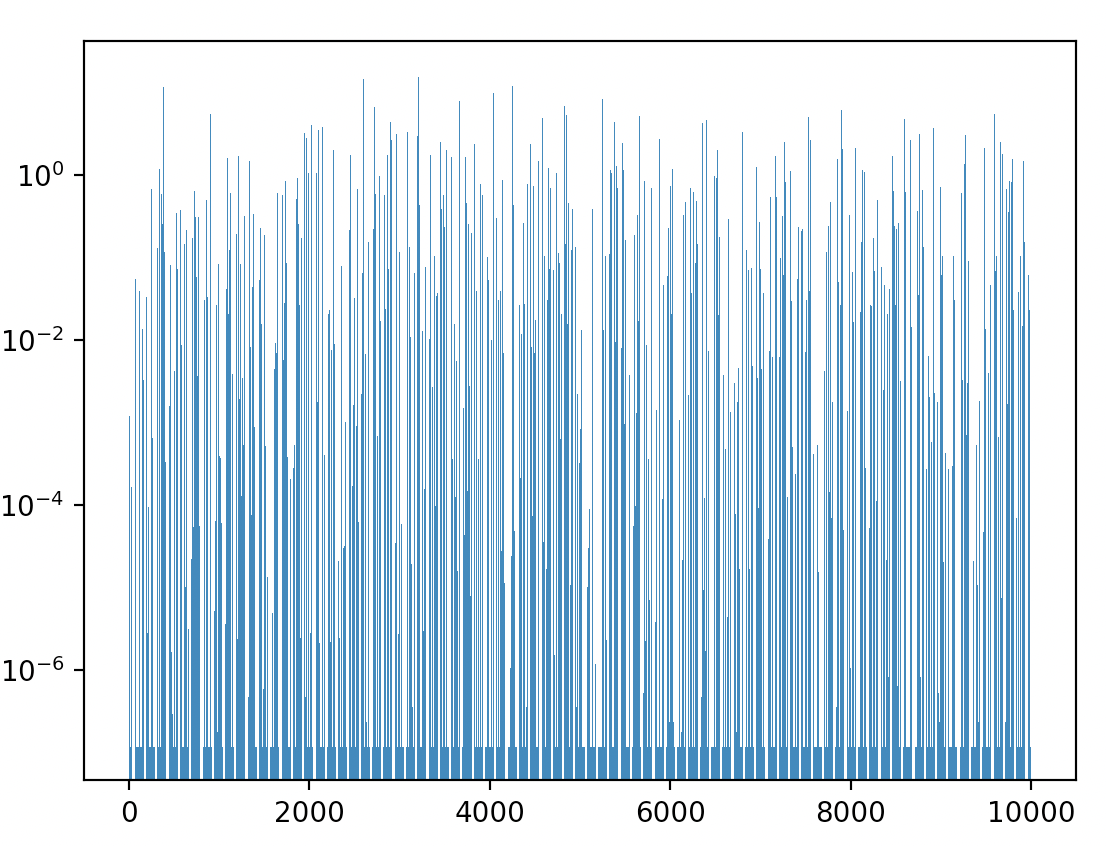}
  \caption{20 epoch}
  \label{fig:sfig1}
\end{subfigure}%
\begin{subfigure}{.5\textwidth}
  \centering
  \includegraphics[width=.9\linewidth]{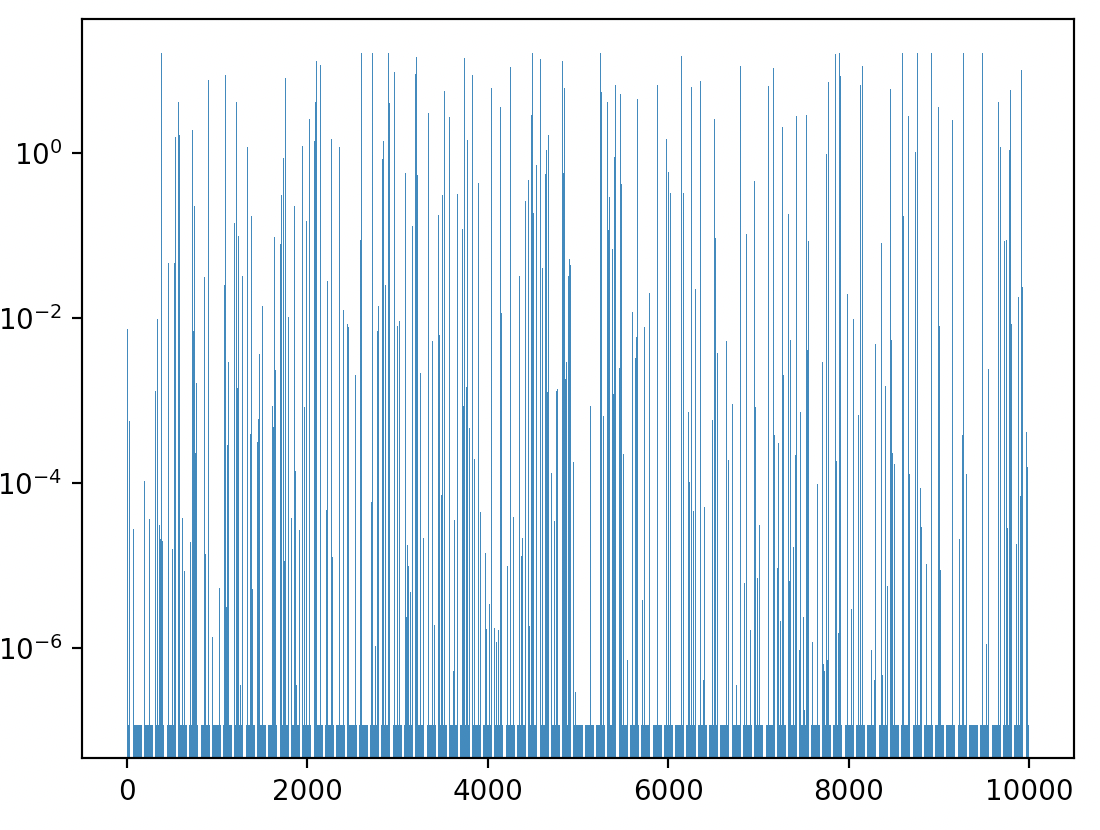}
  \caption{60 epoch}
  \label{fig:sfig2}
\end{subfigure}
\caption[Loss for validation set for two models]{The loss outputs for each example from validation set of \textit{Fashion MNIST} for two sample snapshot models taken respectively from 20 and 60 epoch of the training a single model(one of the models from Fig. \ref{fig:overconfidence}). Please note that the scale is logarithmic.}
\label{fig:bars}
\end{figure}

\begin{figure}[H]
\centering
\includegraphics[scale=0.15]{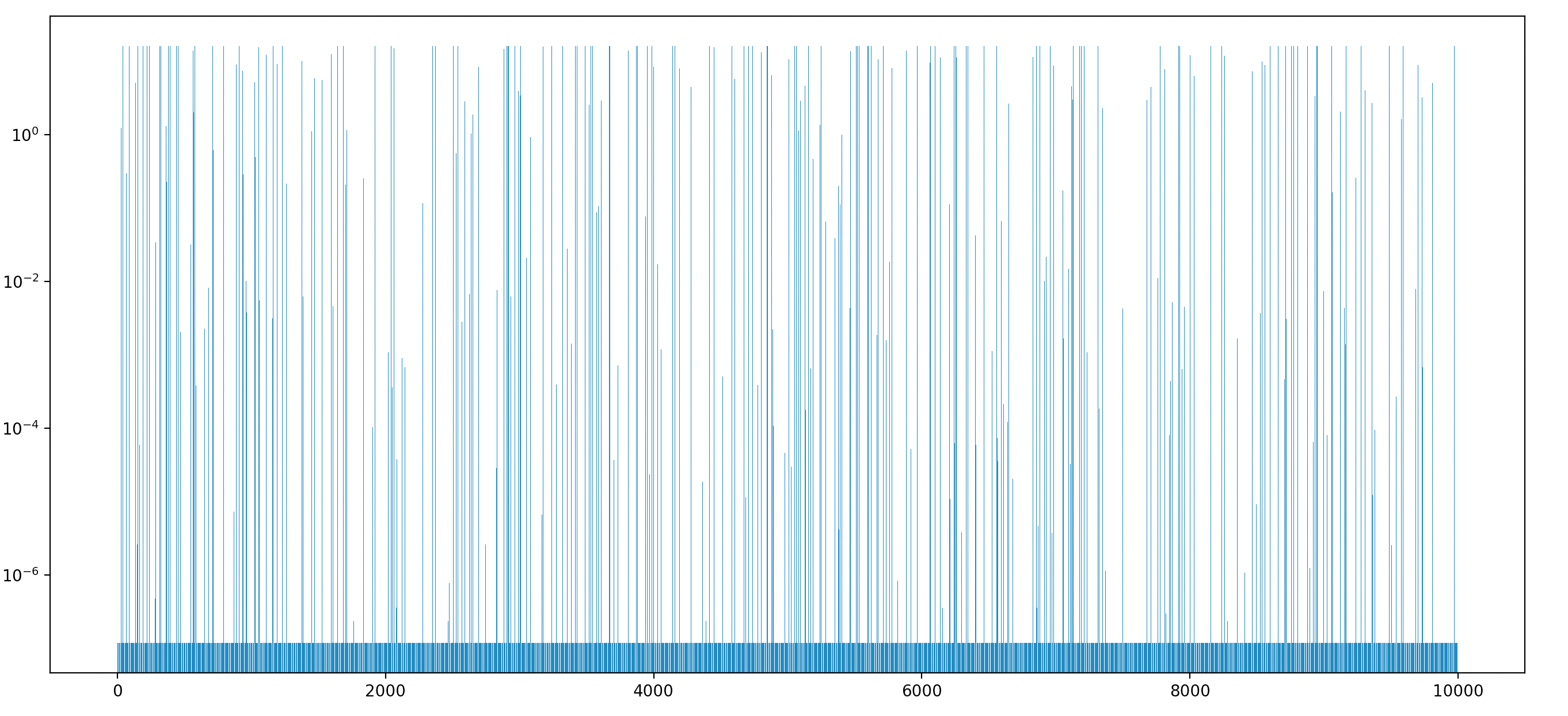}
  \caption[Loss for validation set for an ensemble]{The loss outputs for each example from validation \textit{Fashion MNIST} dataset for \textit{the Supermodel} created from submodels from the 60th epoch of training a single model (one of the models from Fig. \ref{fig:overconfidence}). The scale is logarithmic.}
\label{fig:overconfident_sup}
\end{figure}

\begin{figure}
\begin{subfigure}{.5\textwidth}
  \centering
  \includegraphics[width=1.0\linewidth]{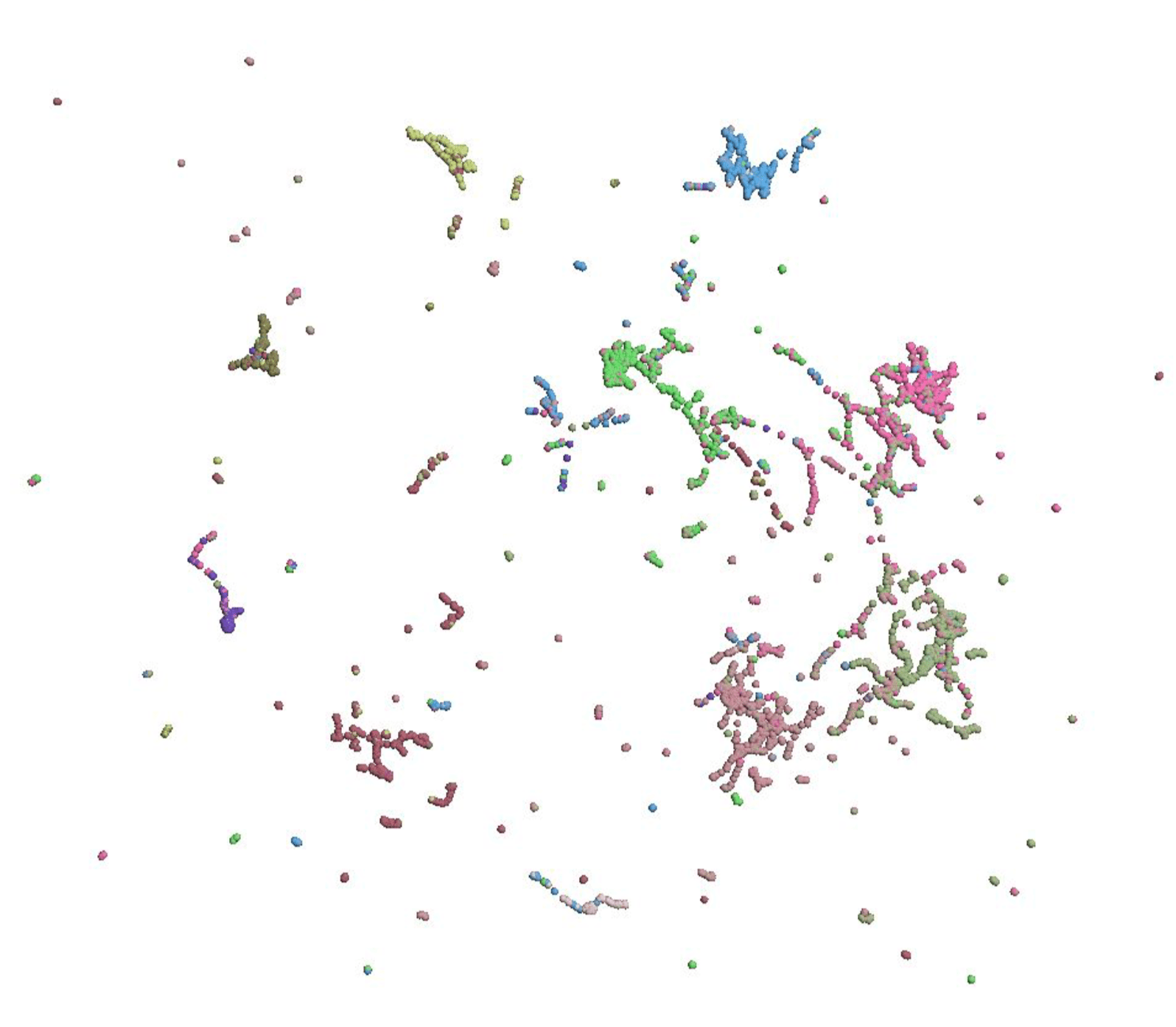}
  \caption{20 epoch. Acc: 0.8954.\break 5-nn acc: 0.846}
  \label{fig:sfig1}
\end{subfigure}%
\begin{subfigure}{.5\textwidth}
  \centering
  \includegraphics[width=1.0\linewidth]{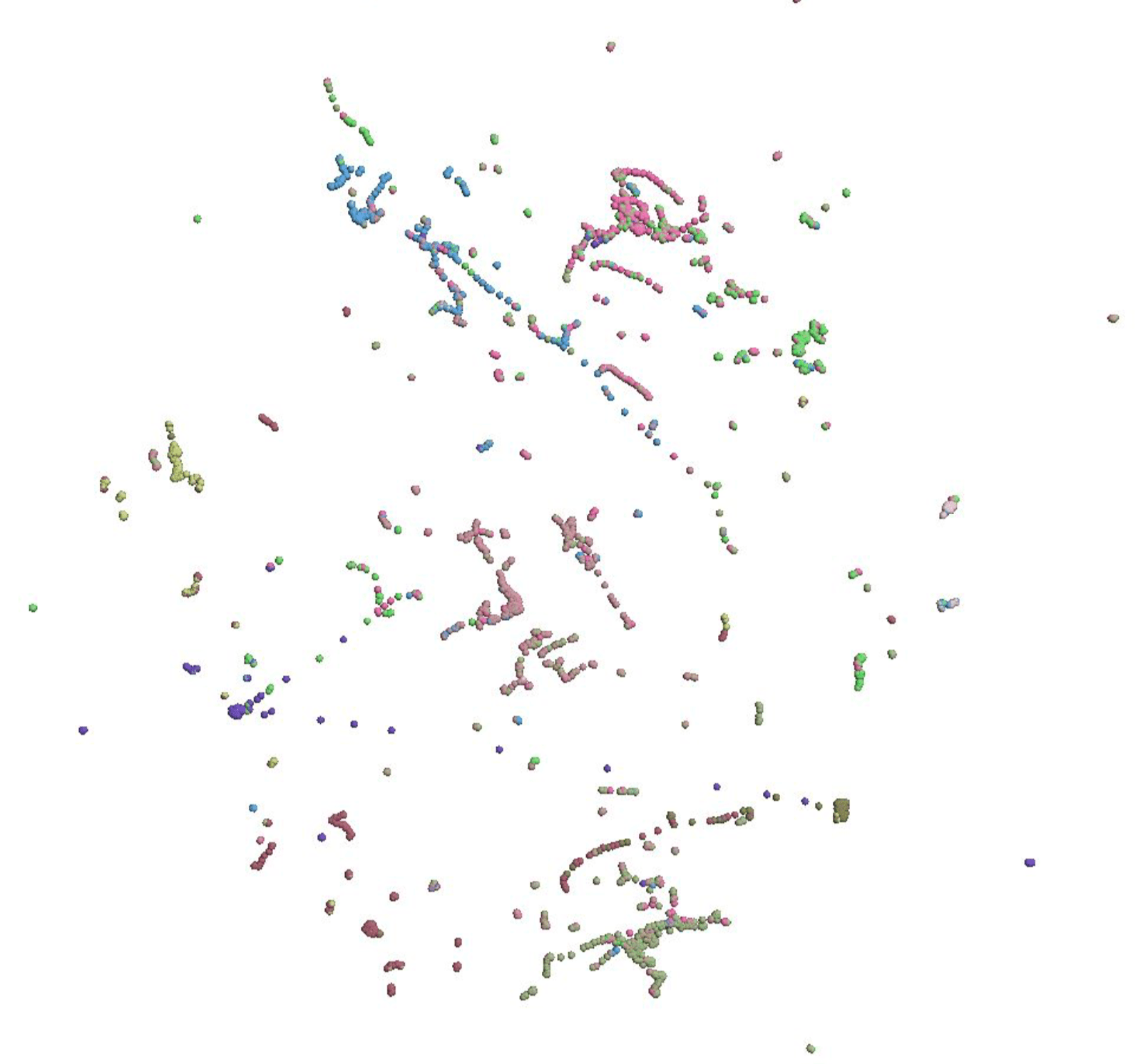}
  \caption{60 epoch. Acc: 0.886.\break 5-nn acc: 0.829}
  \label{fig:sfig2}
\end{subfigure}
\caption[Low and high loss vizualized]{Visualization of the predictions for each example from the validation set of \textit{Fashion MNIST} for two sample snapshot models taken respectively from 20 and 60 epoch of the training a single model(Fig. \ref{fig:overconfidence}). The 5-nn metric was equal 0.846 for 20th epoch and 0.829 for 60th epoch.}
\label{fig:loss_outputs}
\end{figure}

\begin{figure}
\begin{subfigure}{.5\textwidth}
  \centering
  \includegraphics[width=0.8\linewidth]{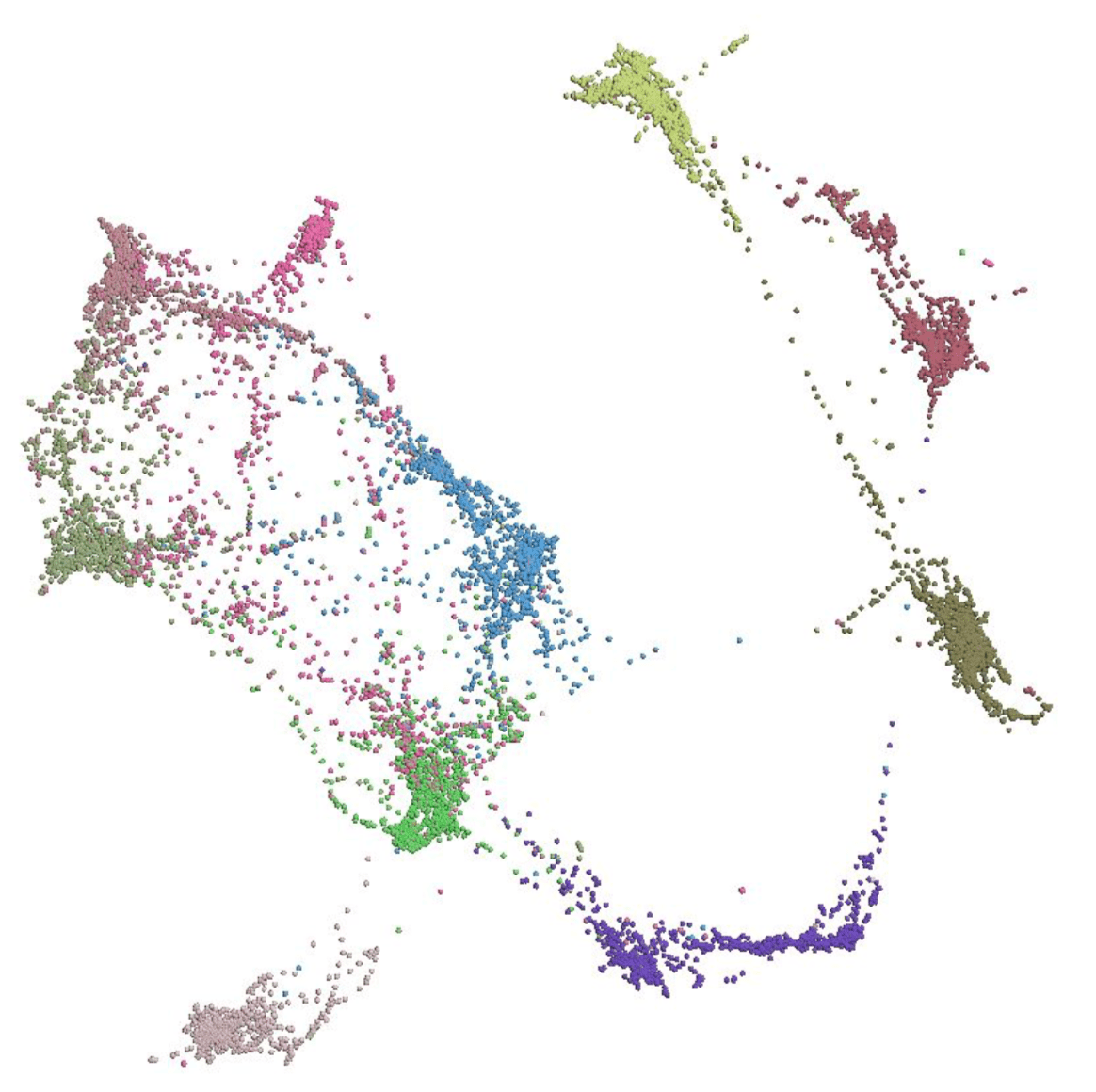}
  \caption{20 epoch. 5-nn acc: 0.866}
  \label{fig:sfig1}
\end{subfigure}%
\begin{subfigure}{.5\textwidth}
  \centering
  \includegraphics[width=0.8\linewidth]{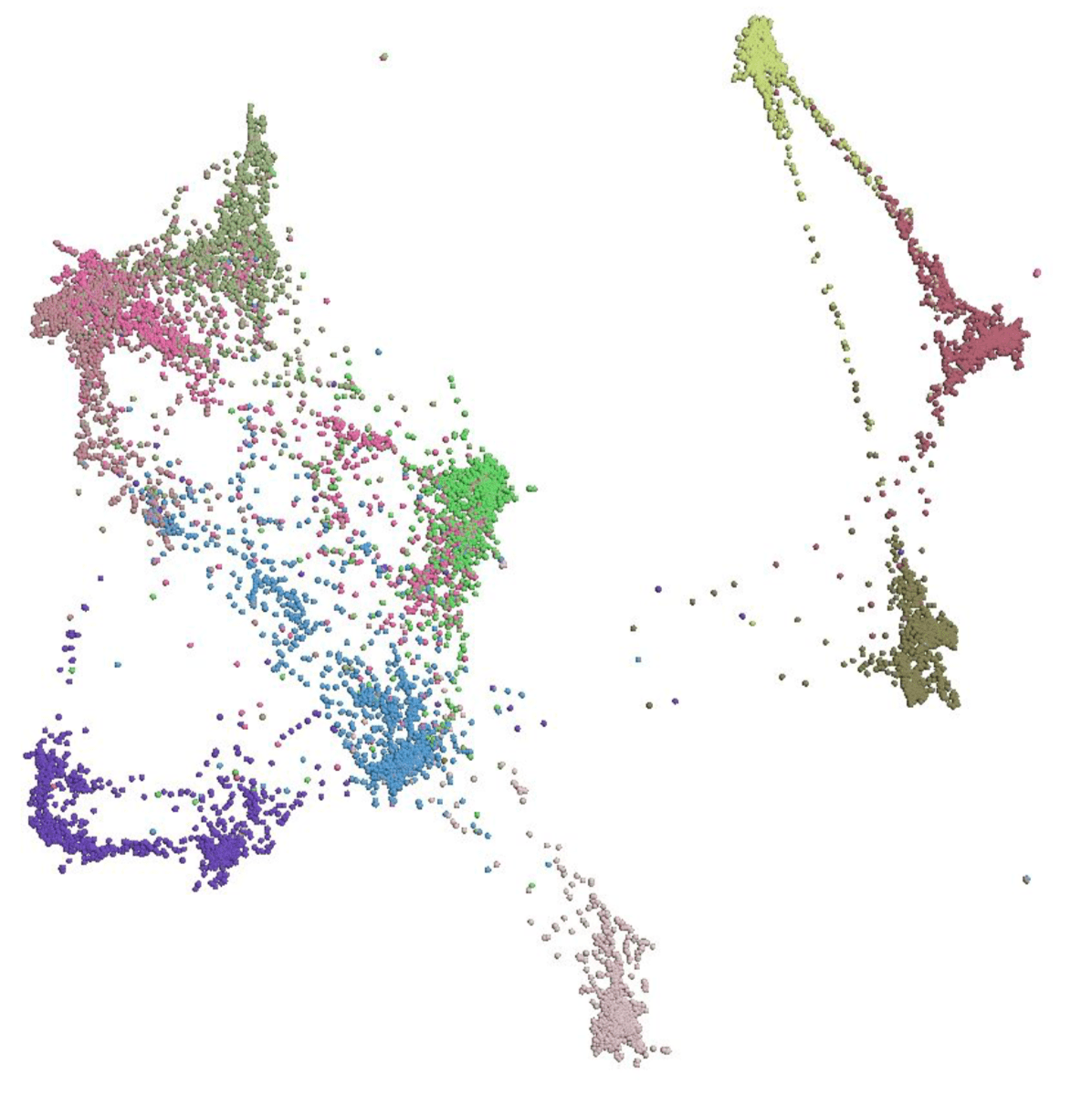}
  \caption{60 epoch. 5-nn acc: 0.867}
  \label{fig:sfig2}
\end{subfigure}
\caption[Low and high loss visualized from penultimate layer]{Visualization of the outputs from the penultimate layer for each example from the validation set of \textit{Fashion MNIST} for two sample snapshot models taken respectively from 20 and 60 epoch of the training a single model(Fig. \ref{fig:overconfidence}). Surprisingly, the 5-nn metric was similar in both cases, equal 0.866 for 20th epoch and 0.867 for 60th epoch.}
\label{fig:loss_outputs2}
\end{figure}

In Fig. \ref{fig:bars} it is visible that the 20th epoch's model has more balanced predictions with less variance. The latter, more overconfident model has outputs that are more close to discrete guesses. Thus, the 60th epoch's predictions are either very precise or completely wrong. In Figures \ref{fig:loss_outputs} and \ref{fig:loss_outputs2} we visualized\cite{embedding}, \cite{embedding2} the outputs for the whole test set for the same two models, from last and penultimate layer respectively. Whereas the 5-nn metric for last layer embedding was bigger for the model with lower loss(0.846 and 0.829), the outputs from penultimate layers had a similar metric, equal ~0.866. This result suggests that the high error of the 60th epoch's model is generated mostly by its final layer. It explains why \textit{SuperModel} for 60th epoch models is not worse than 20th epoch models; when we build an ensemble, we forget the old coefficients of the last layers for each submodel and learn new ones. Analyzing the 5-nn metric from Fig. \ref{fig:loss_outputs2} further, we conclude that model from 60th epoch with re-trained last layer should end up with slightly better accuracy. We confirm such results in Table \ref{tab:last_layer_trained}.
The loss lines from \ref{fig:overconfidence} suggests that the overconfidence of submodels implies higher overconfidence of the ensemble as well. To demonstrate that, Fig. \ref{fig:overconfident_sup} presents loss for each validation example of the \textit{Supermodel} created from 60th epoch of the submodels' training. The bars are either very short or long, which means almost "black or white" guesses.
We can conclude that when choosing the submodels for an ensemble, there is a trade-off between the reliability and accuracy of the \textit{Supermodel's} predictions.

\begin{table}[H]
\centering
\begin{tabular}{|l|l|l|l|l|}
\hline
\multirow{2}{*}{}       & \multicolumn{2}{l|}{\textit{before training last layer}} & \multicolumn{2}{l|}{\textit{after training last layer}} \\ \cline{2-5} 
                        & \textit{val\_acc}          & \textit{val\_loss}          & \textit{val\_acc}          & \textit{val\_loss}         \\ \hline
\textit{epoch 20 model} & 0.8954                     & 0.396                       & 0.8961                     & 0.3765                     \\ \hline
\textit{epoch 60 model} & 0.886                      & 0.682                       & 0.8994                     & 0.4175                     \\ \hline
\end{tabular}
\caption[Last layer trained for two models]{We trained only last layer for two models; from 20th and 60th epoch of the training. We randomly initialized the weights for last layer and used L2 regularization with \textit{alfa} equal 0.01 during the training.}
\label{tab:last_layer_trained}
\end{table}

\subsection{Benefits of Supermodeling}

Let us assume that we have trained the model already, but we would like to reduce the number of parameters without decreasing its final score. The motivation behind this approach might be that the final model is supposed to be uploaded on an embedded device(limited disk space), or we wish to speed up the prediction process. The assumption is that we are still in the process of fine-tuning the architecture.
There are existing methods \cite{pruning} proposed for a neural network compression; however, all of them consider that we are operating on the weights of already pre-trained models. In this paragraph, we try to answer the question, whether \textit{Supermodeling} could be used to compress the architecture \textit{before} the training. 

In Table \ref{tab:compress}, we have compared the performance of \textit{SuperModel} with the normal network that has the same size in terms of weights' space. The results demonstrate that the ensemble achieves better results, starting from a certain size of the model. We can conclude that after reaching some threshold, we can have a higher gain with \textit{Supermodeling} approach for \textit{shallow} networks. However, it may be still possible that properly tuned hyperparameters of a single network may produce better results. Especially for bigger models, in order to prevent overfitting the regularization methods like L2 norm or \textit{dropout}\cite{dropout} should be used. After playing a bit with hyperparameters and extending the training time significantly, the best score that we were able to achieve for \textit{Fashion MNIST} dataset for a single network was 
90.8\%(480, 1120, 1120 and 10 neurons), whereas \textit{SuperModel} reached 91.36 \% (6 subnetworks, 200, 466, 466 and 10 neurons each). 
We have performed similar experiments on \textit{TNG} dataset in Table \ref{tab:ngcompress}.
We have trained the classical models for up to 200 epochs and captured the best score. Having in mind that \textit{dropout} achieves similar performance as L2 norm\cite{dropout_as_ensemble}, we compared ensembling only with \textit{dropout}
For each size, we chose the most optimal model's configuration, based on the manipulation of the \textit{dropout} rate in each layer. The depth of networks was always 3 levels of neurons. Based on high loss, it looks like the single network was reaching its best score being overfitted. In this experiment, the \textit{SuperModel}'s weights were rescaled before training, because it reduces the loss due to results from \textit{Loss function dilema} paragraph. Especially, for the models of bigger sizes, we have received the highest score of around 84\% when the \textit{dropout} rate was large, around 0.7. We would like to recall that one of the explanations why using the \textit{dropout} technique works, is that it may be perceived as an aggregation of the exponential number of ensembles\cite{dropout},\cite{adaptive_drop} in a single model, however, trained independently only one epoch. Thus, the \textit{dropout} gains from averaging properties of the ensembles, which reduce the variance and prevents from overfitting. Significant difference between these two concepts is the measure of correlation between neurons. The neurons from two submodels in the ensemble are disconnected, whereas \textit{dropout} "generates" tightly coupled submodels, with shared weights. This is just a specific point of view on \textit{dropout}, and for further reading, please refer to the literature\cite{understand_dropout}.
Analyzing the results from Tables \ref{tab:compress} and \ref{tab:ngcompress}, it appears that the ensemble learning achieves better results, notably for bigger models. We hypothesize that starting from a certain level of model's complexity, \textit{the Supermodeling} gives a better correlation degree between the feature detector units(understood as neurons or group of neurons) than hypothetical ensembles "created" by standard \textit{dropout} approach\cite{error_correlation}. In less correlated ensembles, the units are more independent and do not share that much information. This may lead to the conclusion that \textit{the Supermodeling} could be used to \textit{regularize} very shallow and wide networks by sort of structure modification during training. This would partially explain why convolutional networks work so well as the gain come from separate autonomous filters. However, in order to prove such the hypothesis, a separate study would be necessary, and our conclusions are rather intuitive guesses than formally proved theorems.
We did not use the word "compression" here on purpose - it would imply that the classical network always reproduces the score of the \textit{Supermodel}. However, that was not a case, and we did not manage to achieve the same result of 85\% on \textit{TNG} dataset and 91\% on \textit{F-MNIST} with the classical model of any size. Obviously, that may be the matter of perfectly chosen hyperparameters and regularization. Therefore, the \textit{Supermodeling} seems to be a more natural and quicker way for trivially boosting performance without spending much effort on the architecture's fine-tuning.

\begin{table}[H]
\centering
\begin{adjustbox}{width=1.0 \textwidth}
\begin{tabular}{|l|l|l|l|l|l|l|l|}
\hline
number of parameters          & \multicolumn{1}{c|}{53 k} & \multicolumn{1}{c|}{115 k} & \multicolumn{1}{c|}{360 k}           & 687 k           & 1,257 k         & 1,781 k         & 2,837 k         \\ \hline
Classical MLP                 & 0.8826                    & 0.8928                     & \multicolumn{1}{r|}{\textbf{0.9023}} & \textbf{0.9057} & 0.9049          & 0.9074          & 0.9054          \\ \hline
\textit{SuperModel} & 0.8559                    & 0.8724                     & 0.8941                               & 0.9025          & \textbf{0.9071} & \textbf{0.9081} & \textbf{0.9113} \\ \hline
\textit{SuperModel whole trained} & \textbf{0.8845}           & \textbf{0.8947}            & 0.9018                               & 0.9038          & 0.9025          & 0.9016          & 0.9039          \\ \hline
\end{tabular}
\end{adjustbox}
\caption[Classical MLP vs SuperModel for \textit{Fashion MNIST}]{
Validation accuracy for 7 different sizes of classical networks and corresponding \textit{SuperModels} for 100 epochs of training on \textit{Fashion MNIST} dataset.
The base model had 53k trainable parameters and consisted of four layers; 45, 105, 105, and 10 neurons. We set \textit{dropout} rate to 0.2 between layers.
We scaled the model by multiplying the number of neurons in each layer by different factors, which produced seven different architectures(each column).
Then, each model was compared with two types of \textit{SuperModels} (6 submodels) that had a similar number of parameters and the same 4-layers depth. The score of the \textit{Supermodel} trained from completely random initialization proves that it is not that good as when we trained subnetworks separately (already shown in the Fig. \ref{fig:supermodels_compare});
}
\label{tab:compress}
\end{table}

\begin{table}[H]
\centering
\begin{adjustbox}{width=1.0 \textwidth}
\begin{tabular}{|c|l|l|l|l|l|l|l|l|}
\hline
\multicolumn{1}{|l|}{{\color[HTML]{000000} number of parameters}} & {\color[HTML]{656565} }          & {\color[HTML]{000000} 300k}            & {\color[HTML]{000000} 600k}            & {\color[HTML]{000000} 1,15 kk}         & {\color[HTML]{000000} 1,81 kk}         & {\color[HTML]{000000} 3,09 kk}         & {\color[HTML]{000000} 5,15 kk}         & {\color[HTML]{000000} 7,3 kk}          \\ \hline
{\color[HTML]{000000} }                                           & {\color[HTML]{000000} val\_acc}  & {\color[HTML]{000000} \textbf{0.8129}} & {\color[HTML]{656565} 0.8185}          & {\color[HTML]{656565} 0.8343}          & {\color[HTML]{656565} 0.8445}          & {\color[HTML]{656565} 0.8406}          & {\color[HTML]{656565} 0.8437}          & {\color[HTML]{656565} 0.8426}          \\ \cline{2-9} 
\multirow{-2}{*}{{\color[HTML]{000000}\textit{ Single model}}}             & {\color[HTML]{000000} val\_loss} & {\color[HTML]{656565} 0.7637}          & {\color[HTML]{656565} 0.7144}          & {\color[HTML]{656565} 0.7051}          & {\color[HTML]{656565} 1.0552}          & {\color[HTML]{656565} 0.9098}          & {\color[HTML]{656565} 1.6795}          & {\color[HTML]{656565} 1.5173}          \\ \hline
{\color[HTML]{000000} }                                           & {\color[HTML]{000000} val\_acc}  & {\color[HTML]{000000} 0.8016}          & {\color[HTML]{000000} \textbf{0.8261}} & {\color[HTML]{000000} \textbf{0.8348   }} & {\color[HTML]{680100} \textbf{0.8502}} & {\color[HTML]{000000} \textbf{0.8465}} & {\color[HTML]{000000} \textbf{0.8504}} & {\color[HTML]{000000} \textbf{0.8470}} \\ \cline{2-9} 
\multirow{-2}{*}{{\color[HTML]{000000} \textit{SuperModel}}}                 & {\color[HTML]{000000} val\_loss} & {\color[HTML]{656565} 0.7585}          & {\color[HTML]{656565} 0.7686}          & {\color[HTML]{656565} 0.6600}          & {\color[HTML]{656565} 0.6649}          & {\color[HTML]{656565} 0.9174}          & {\color[HTML]{656565} 0.8600}          & {\color[HTML]{656565} 0.7480}          \\ \hline
\end{tabular}
\end{adjustbox}
\caption[Classical MLP vs SuperModel for \textit{TNG}]{
Validation accuracy for 7 different sizes of classical networks and corresponding \textit{SuperModels} training on \textit{TNG} dataset. }
\label{tab:ngcompress}
\end{table}


\subsection{Summary}

Summarising what we presented in this chapter:
\begin{enumerate}
\item In the "\textit{Last layers training}" section, we have introduced a novel method for boosting the accuracy of a single NN model.
\item Then, the "\textit{Partitioning}" section proposes the specific ensembling method that improves the classification performance by combining multiple submodels of the partitioned network
\item In the section "\textit{Loss function dilemma}", we explain why increasing validation accuracy does not always come together with decreasing loss for cross-entropy function. 
\item In the section "\textit{Best submodels}", we presented that slightly overconfident networks create compositions that are more accurate and confident.
\item "\textit{Benefits of Supermodeling}" points out that the efficiency of the ensemble method differs from the classical models depending on the scale of parameters.

\end{enumerate}

Let us recall the Table\ref{tab:cutmodel} from the introduction of this chapter.
We concluded that the ensemble achieves a higher score than a single network, but it is more time-consuming. However, to prevent overfitting, the training of the subnetworks was terminated when the validation loss stopped to decrease.
At this last stage of our research around partitioning fully-connected networks we would like to compare the \textit{CPU time} of the \textit{Supermodel} and an individual model until both reached the same validation accuracy, regardless of the architecture and size. We fine-tuned single model and an ensemble in order to achieve 90\% precision on \textit{Fashion MNIST} and 84\% on \textit{TNG}. We present the results in Tables \ref{tab:bestscore} and  \ref{tab:bestscore2}. We conclude that ensembling can be used for saving CPU resources as well. However, we need to keep in mind that all those experiments were done on 
MLP networks, and convolutional architectures could achieve greater scores on the mentioned datasets — results from convolutional types of networks we present in the next chapter.

\begin{table}[H]
\centering
\begin{tabular}{|l|l|l|l|}
\hline
           & val acc & val loss & time              \\ \hline
single net & 0.9     & 0.3228   & \textbf{00:08:36} \\ \hline
supermodel & 0.9     & 0.3198   & \textbf{00:04:59} \\ \hline
\end{tabular}
\caption[CPU time required for achieving 90\% on \textit{Fashion MNIST}]{ We compare CPU time required for reaching 90\% accuracy on \textit{Fashion MNIST} for a best individual model and an ensemble. The \textit{Supermodel} consisted of 3 subnetworks and was significantly faster.
}
\label{tab:bestscore}
\end{table}

\begin{table}[H]
\centering
\begin{tabular}{|l|l|l|l|}
\hline
           & val acc & val loss & time              \\ \hline
single net & 0.84    & 1.1336   & \textbf{00:03:00} \\ \hline
supermodel & 0.84    & 0.6508   & \textbf{00:01:13} \\ \hline
\end{tabular}
\caption[CPU time required for achieving 84\% on \textit{TNG}]{ We compare CPU time required for reaching 84\% accuracy on \textit{TNG} for a best individual model and an ensemble.
}
\label{tab:bestscore2}
\end{table}
\section{Supermodeling of deep architectures} 
\label{chap:general-concept}

\subsection{State of art results}

\begin{figure}[H]
\centering
\includegraphics[scale=0.2]{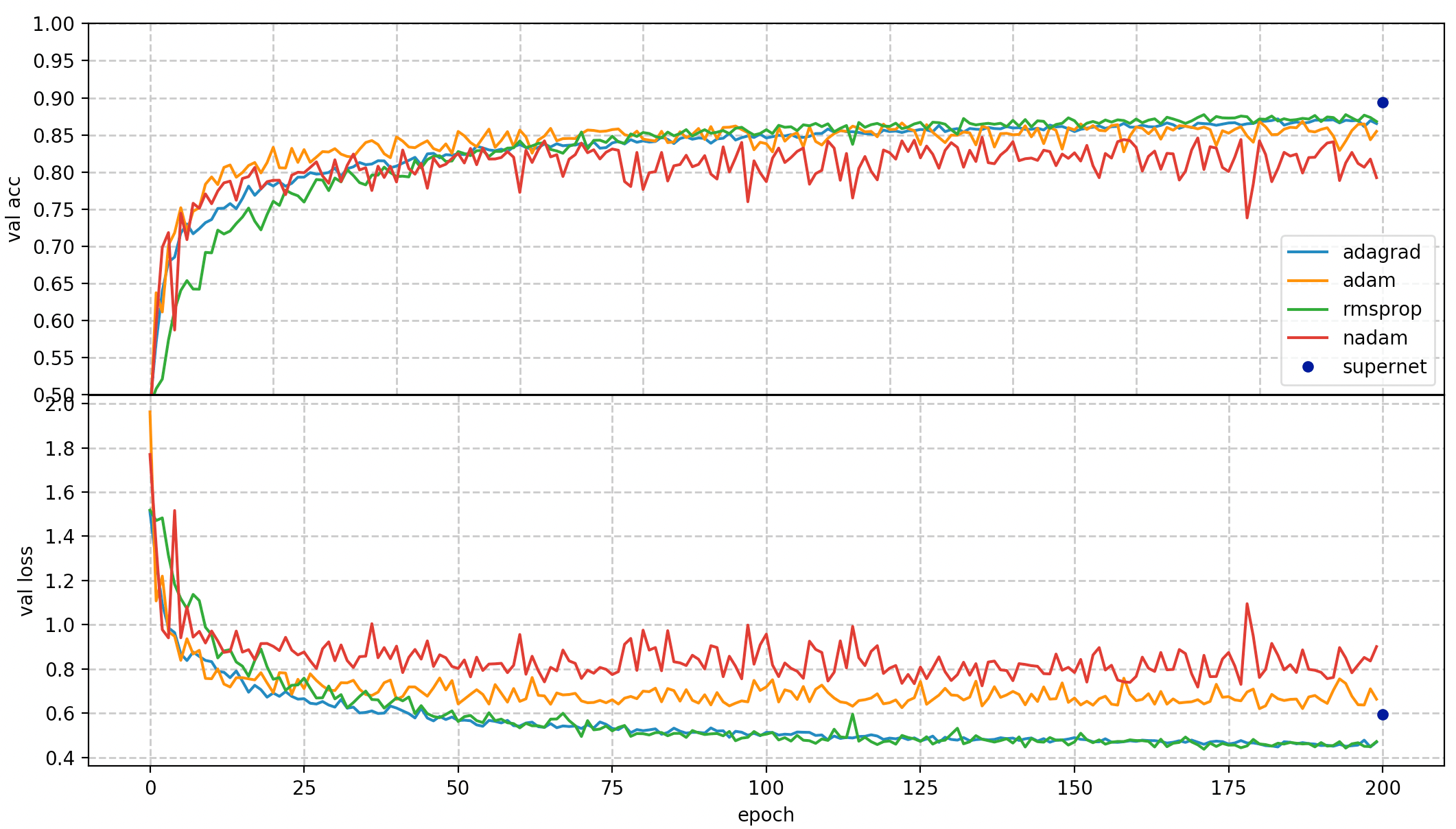}
  \caption[\textit{SuperNet} with different optimizers]{
  Validation accuracy and loss of training four models from Table \ref{tab:mymodel}. Blue dot indicates the accuracy achieved by \textit{SuperNet} created from those models.
 We do not present \textit{SuperNet's} training time because it is significantly shorter than subnetworks, as we train only the last, fully-connected layer. Usually, it takes less than ten epochs, and it is overfitting sensitive.}
\label{fig:supernet_ex1}
\end{figure}

\begin{figure}[H]
\centering
\includegraphics[scale=0.18]{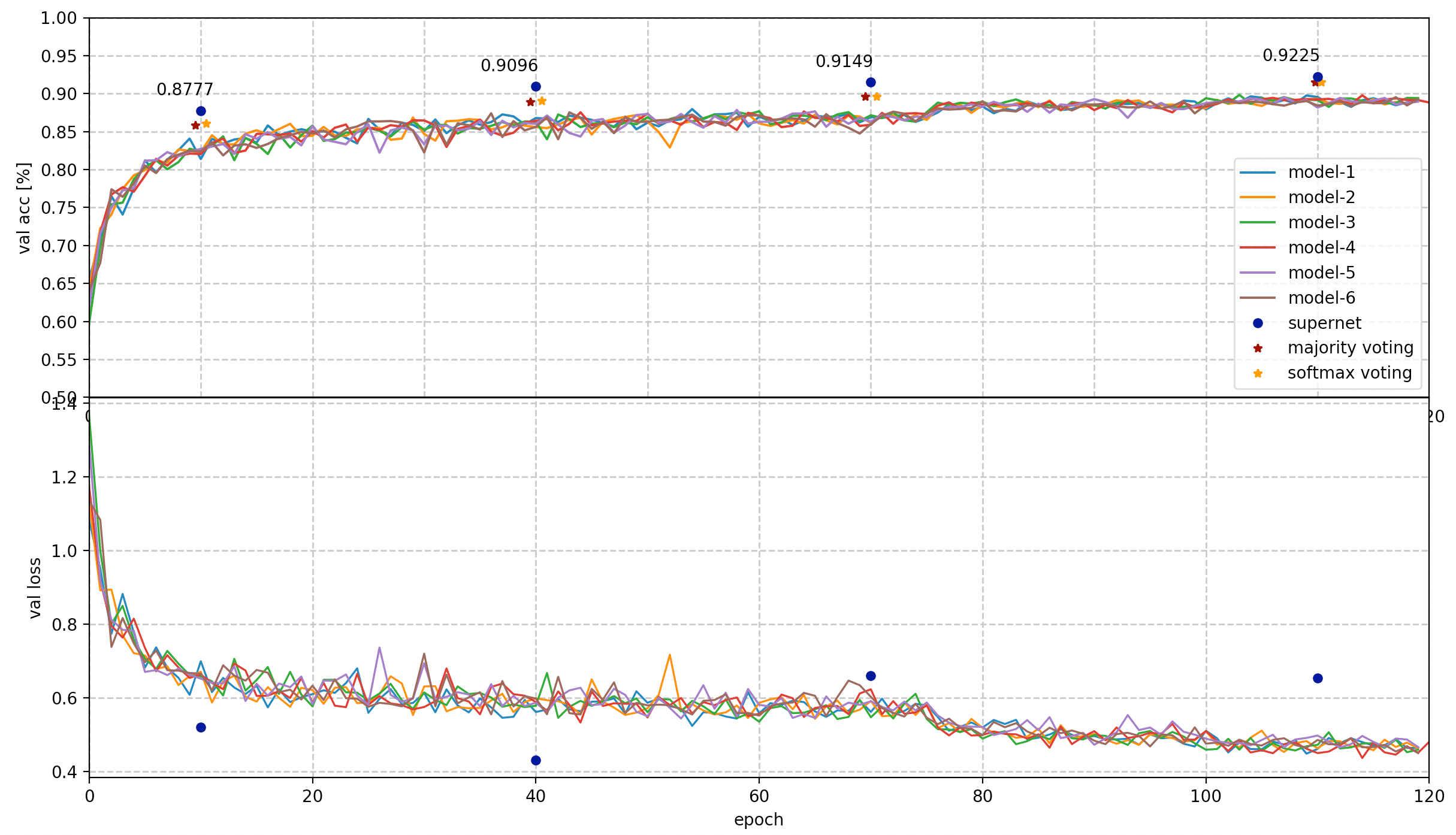}
  \caption[\textit{SuperNet} on \textit{CIFAR10}]{\textit{SuperNet} based on six models, saved in 10,40, 70 and 110 epoch of training on \textit{CIFAR10} dataset for a model from Table \ref{tab:mymodel}. The same optimizer \textit{Adam} was used for each network.
  The result is compared to \textit{Majority voting} method and \textit{Softmax voting}. \textit{Softmax voting} is the voting based on the sum of probabilities outputs from the subnetworks. The results are from different training than values presented in Table \ref{tab:accsubs}.}
\label{fig:supernet_ex2}
\end{figure}

\begin{figure}[H]
\centering
\includegraphics[scale=0.18]{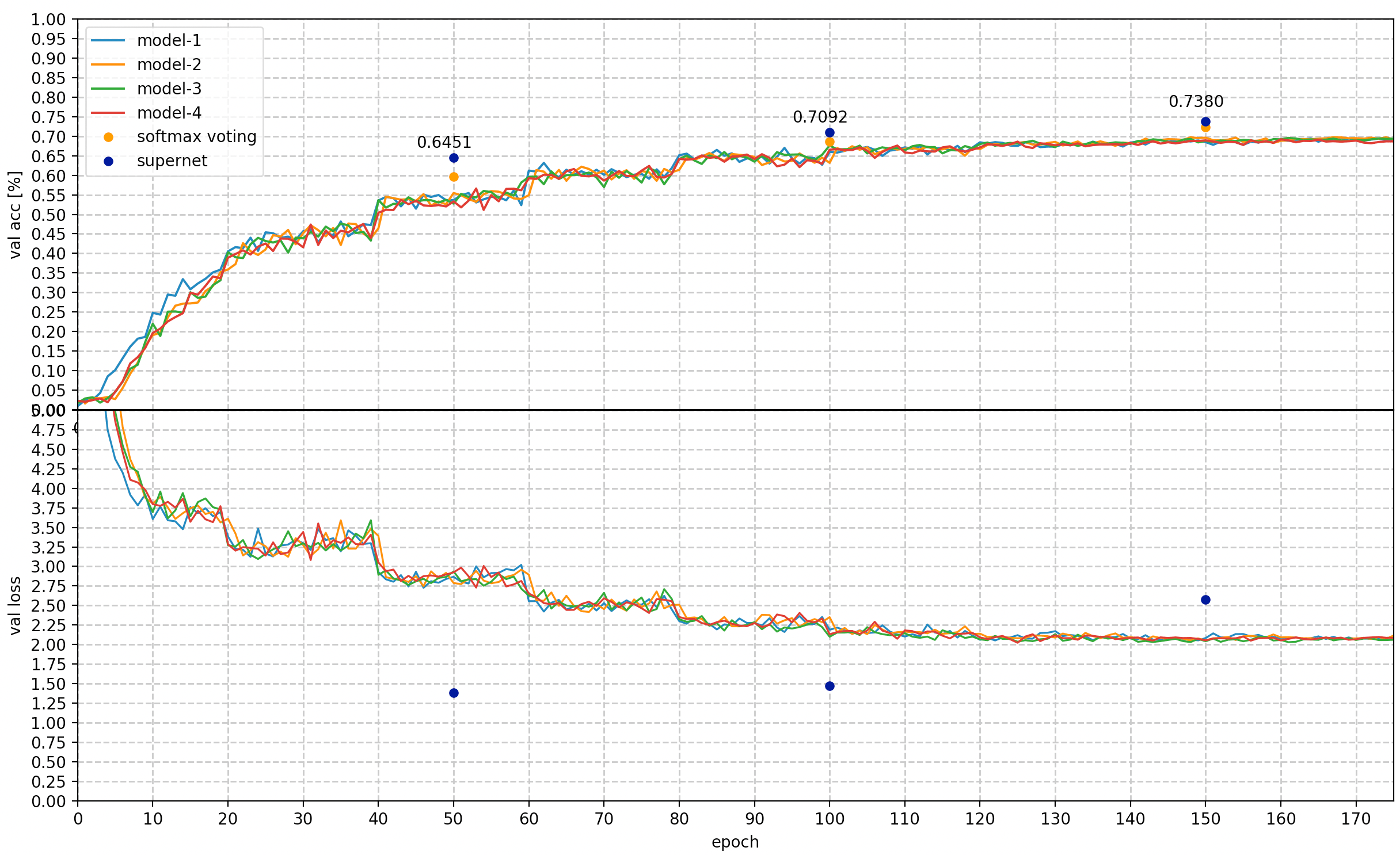}
  \caption[\textit{SuperNet} on \textit{CIFAR100}]{\textit{SuperNet} based on six models, saved in 50,100, and 150 epoch of training on \textit{CIFAR100} dataset for VGG\cite{vgg} model.
  The result is compared to \textit{Majority voting} method and \textit{Softmax voting}.}
\label{fig:supernet_ex3}
\end{figure}

Setting aside the considerations around MLP models, we have experimented with \textit{Supermodeling} of deep architectures. For that part of experiments, we used \textit{CIFAR10} and \textit{CIFAR100} datasets\cite{cifar} and model presented in Table \ref{tab:mymodel}. Fig. \ref{fig:supernet_ex1} demonstrates the results of validation accuracy and loss for four independently trained NN models(we call it here, in contrast to the previous "partition" \textit{SuperModels}, \textit{SuperNets}). Each network was trained up to 200 epochs with varoius optimizers: \textit{adagrad, rmsprop, nadam, and adam}. This is important because as shown in \cite{optimizers}, the quality of trained NN model depends heavily on the optimization method used. The best choice, in turn, depends on both dataset considered and the NN architecture employed.
After the training, we have created the \textit{SuperNet} and additionally fit the last layer.
Figures \ref{fig:supernet_ex2} and \ref{fig:supernet_ex3} present the results of \textit{SuperNet} of six models in different stages of the training for \textit{CIFAR10} and \textit{CIFAR100} datasets respectively. The results present that \textit{SuperNet} wins over the classical voting ensembles.
In Table \ref{tab:accsubs} we have measured the \textit{SuperNet's} scores for different numbers of subnetworks.
Based on Table, we can conclude that the higher number of models implies the better accuracy of their \textit{SuperNet}. 
We additionally implemented the idea of re-training just the last layers for those six submodels and after that combined them into the \textit{SuperNet}. We achieved 92.48\% validation accuracy on the \textit{CIFAR10} dataset. This result reproduces the best score for the year 2015 for that particular dataset\cite{benchmarks}, which is quite impressive as we are considering an ensemble of rather simple convolutional networks.

\begin{table}[]
\centering
\small
\begin{adjustbox}{width=1.0 \textwidth}
\begin{tabular}{@{}|l|l|l|l|l|l|l|@{}}
\toprule
Models      & 6 submodels      & 5 submodels & 4 submodels & 3 submodels & 2 submodels & 1 submodel \\ \midrule
$model_1$ acc & 0.8636          & 0.8636     & 0.8636     & 0.8636     & 0.8636     & 0.8636     \\ \midrule
$model_2$ acc & 0.8619          & 0.8619     & 0.8619     & 0.8619     & 0.8619     & -          \\ \midrule
$model_3$ acc & 0.8548          & 0.8548     & 0.8548     & 0.8548     & -          & -          \\ \midrule
$model_4$ acc & 0.8489          & 0.8489     & 0.8489     & -          & -          & -          \\ \midrule
$model_5$ acc & 0.8434          & 0.8434     & -          & -          & -          & -          \\ \midrule
$model_6$ acc & 0.8739          & -          & -          & -          & -          & -          \\ \midrule
SuperNet & \textbf{0.9120} & 0.9104     & 0.9083     & 0.9044     & 0.901      & 0.8894     \\ \midrule
Softmax  & 0.8973          & 0.8904     & 0.8925     & 0.8892     & 0.8832     & 0.8636     \\ \bottomrule
\end{tabular}
\end{adjustbox}
\caption[Different number of submodels for an ensemble]{
Accuracy of \textit{the SuperNet} and \textit{Softmax voting} with respectively 6,5,4,3,2 and 1 subnetworks included for \textit{CIFAR20} dataset and the model from Table \ref{tab:mymodel}. The results show that more pre-trained submodels effect in better \textit{SuperNet's} accuracy.
}
\label{tab:accsubs}
\end{table}

\subsection{Loss surfaces analysis}

We have already proven in Table \ref{tab:accsubs} that we a better improvement for an ensemble occurs when we include a higher number of submodels. However, we know that having multiple pre-trained networks is computationally expensive, as we need to train them independently. Finding diverse networks easily is not a trivial task but already addressed in the literature\cite{m_free}, \cite{garipov}.
If we think about neural network training from a geometrical point of view, the goal is to find a global minimum in the highly dimensional surface of a loss function. The shape of the surface depends on multiple factors like the chosen model, number of trainable parameters, dataset, and more. There are multiple studies around the topic of visualization landscapes\cite{visualise_landscape} of the loss functions and the analyses of them \cite{emp_loss},\cite{surfaces}. 

In my thesis, we would like to put more attention on one - "Loss Surfaces, Connectivity Mode and Fast Ensembling of DNNs"\cite{garipov} by Timur Garipov.
The author discovers the fascinating property of the surfaces of the deep neural networks. His empirical researches present that every two local optima are connected by the simple curves or even triangle polyline over which training and test accuracy are nearly constant. The interesting is the fact that the independent research team\cite{no_barriers} discovered the same property of the multidimensional loss function's surface. Their approach was based on the Automated Nudged Elastic Band algorithm\cite{automated}, which addresses problems related to physics and chemistry.
That leads to the conclusion that is just one pre-trained network could lead, without large computational budget, to an infinite number of other, \textbf{diverse} networks with similar performance. It comes from the assumption that our network is located in some "valley" of low loss error. Moreover, Garipov in \cite{garipov} proposes a unique method of training neural networks, called Fast Geometrical Ensemble(FGE), that finds such models and ensemble those snapshots in order to achieve improved accuracy. Inspired by the geometrical insights, the approach is to adopt a cyclical learning rate that forces the optimizer (e.g., gradient descent) to "travel" along mode connectivity paths. Even though it is not the first introduction of cyclical learning rate in the domain literature(\cite{cyclical}, \cite{m_free}), the author highlights that his method uses a very small cycle compared to other works and finds diverse models relatively faster. 

\begin{figure}[H]
\centering
\includegraphics[scale=0.6]{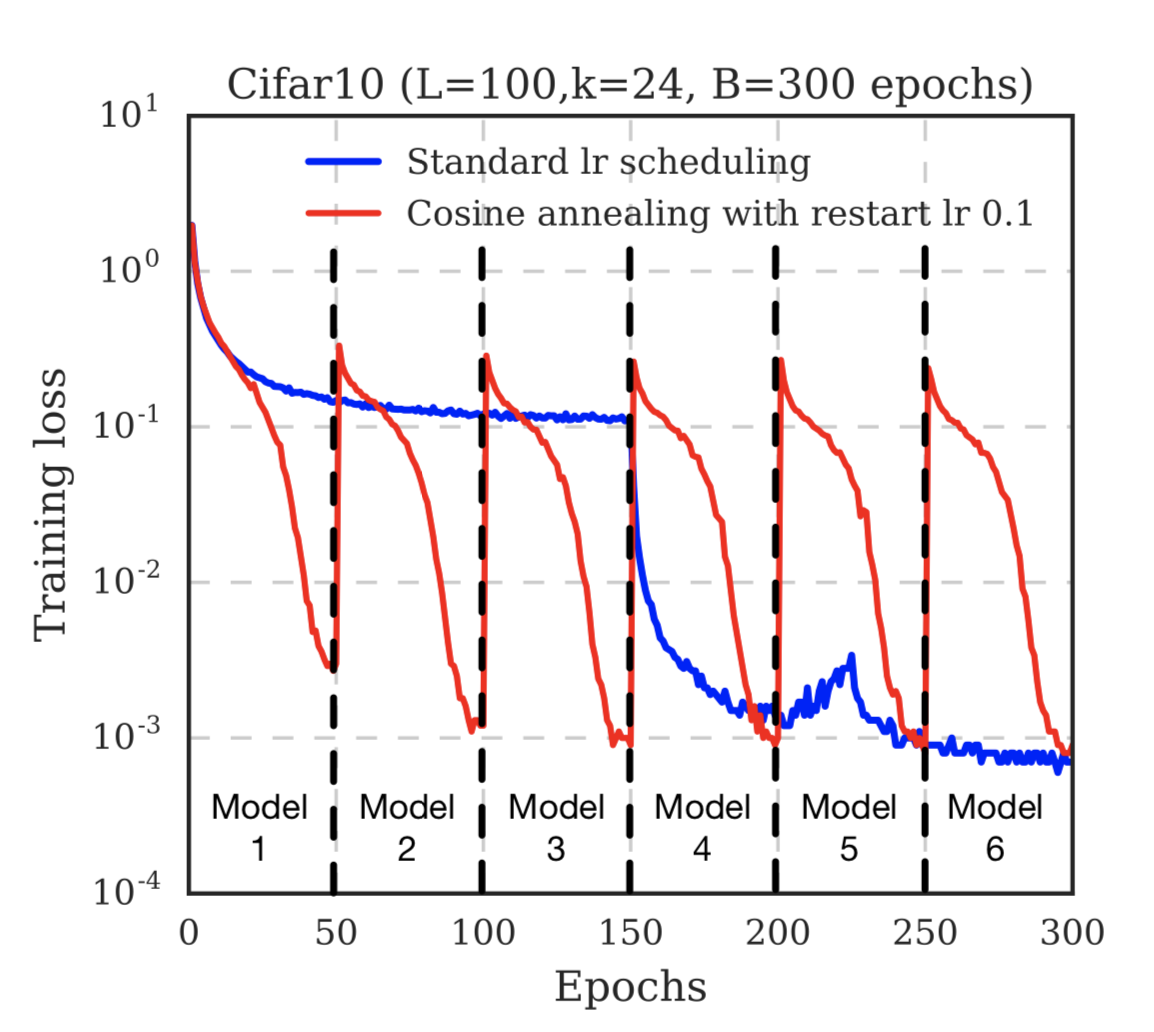}
  \caption[Cyclic learning rate explained]{Training loss of standard learning rate and cyclic learning rate on DenseNet\cite{densenet}, on \textit{CIFAR10}. We took a figure from "Snapshot ensembles: Train 1, Get M for Free"\cite{m_free}}
\label{fig:cyclic}
\end{figure}

The general idea behind such a cyclic learning rate, sometimes called \textit{cosine annealing learning rate}, is to assume that we have a cycle that consists of a constant number of epochs. Each cycle starts with a relatively large learning rate, which is consequently decreased with the training time. At the end of the cycle, it should be comparable to a standard learning rate. Then, one saves the snapshot of the model and starts the process again.
We present in Fig. \ref{fig:cyclic} a comparison of training loss of standard learning rate and cyclic learning rate on DenseNet. In the next paragraph, we are testing such a method for obtaining submodels for our ensemble architecture.

\subsection{Snapshot Supermodeling}
\begin{figure}[H]
\centering
\includegraphics[scale=0.4]{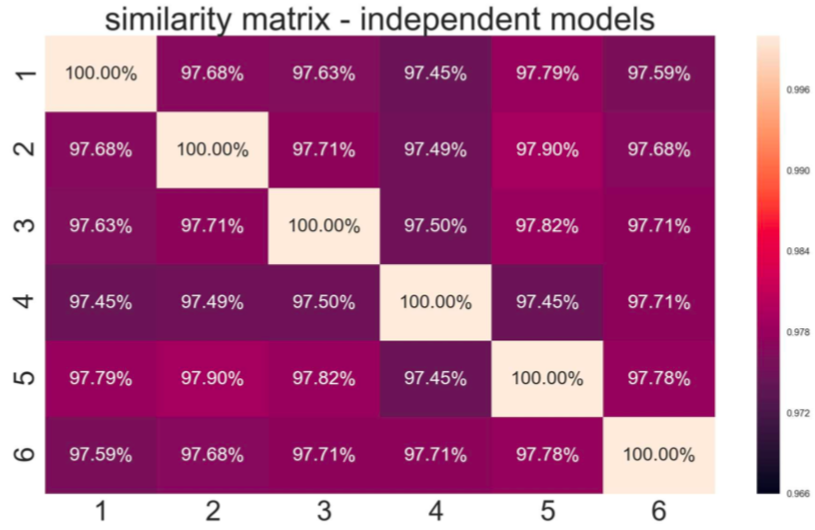}
  \caption[Similarity matrix of independent models]{
  The percentage of the same classifications for six independently trained \textit{DenseNets}.
  }
\label{fig:independent_matrix}
\end{figure}

\begin{figure}[H]
\centering
\includegraphics[scale=0.4]{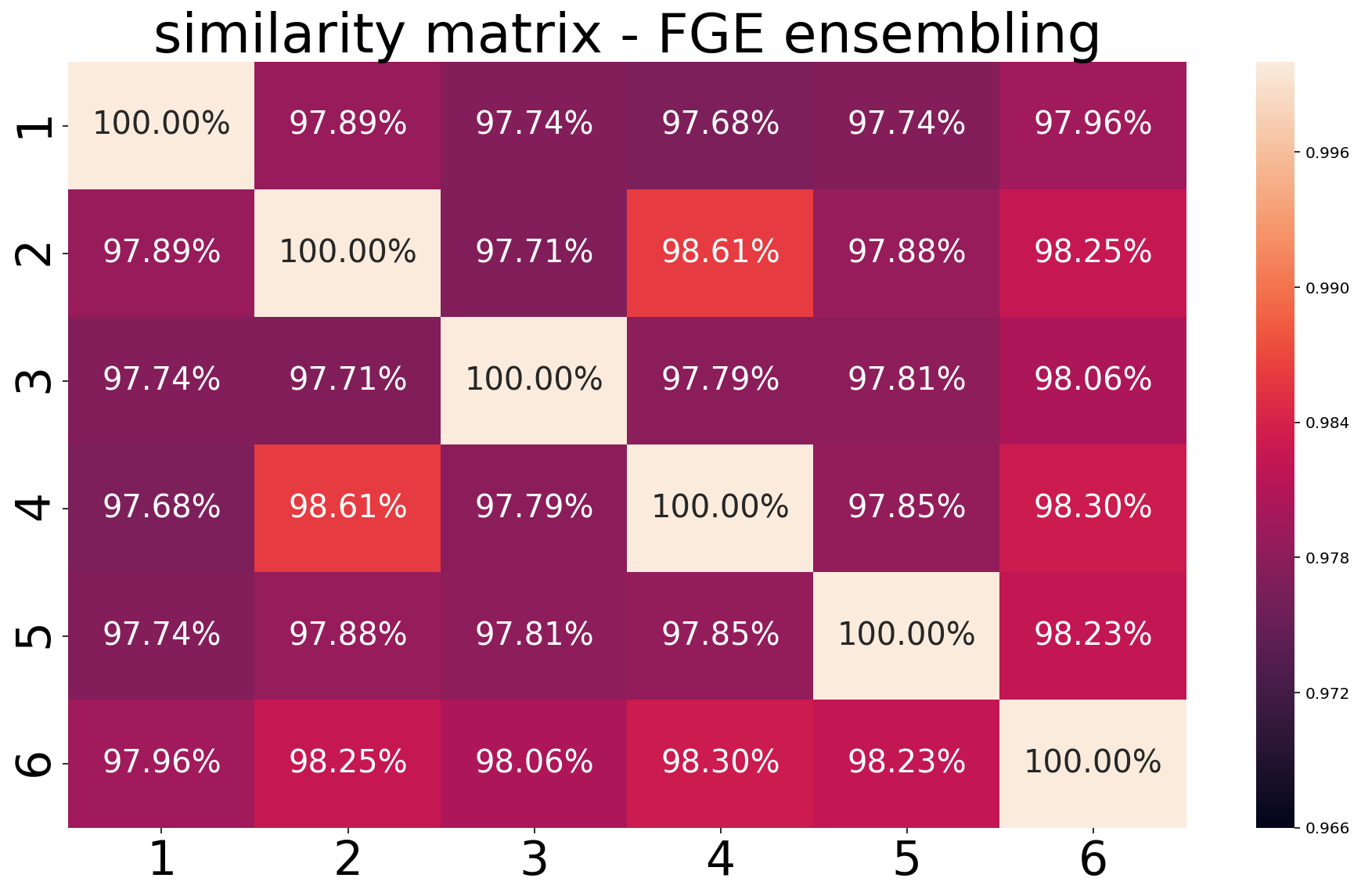}
  \caption[Similarity matrix of FGE models]{
  The percentage of the same classifications for six \textit{DenseNet} models obtained by FGE method.
  }
\label{fig:fge_matrix}
\end{figure}

We compare an accuracy reached by an ensemble that we built from submodels achieved by \textit{FGE}\cite{garipov} method in Table \ref{tab:fge}. We conclude that \textit{FGE} method does not bring diverse models as independent training. However, still, the gain of \textit{Supermodeling} of such the models is notable. 
There are various studies \cite{diversity} around the topic of diversity of models combined into ensembles. In my thesis, we recall the results of Konrad Zuchniak\cite{zuchniak}, which present similarity matrices, measured as percentage of the examples classified to the same class by two models. Figures \ref{fig:independent_matrix} and \ref{fig:fge_matrix} illustrate matrices for independently trained models and \textit{FGE} snapshots, respectively. We observe that the \textit{FGE} snapshots are less diverse, which explain why their \textit{Supermodeling} does not perform that well as in comparison to independently trained models.
As of the last experiment, we combined all the optimization methods mentioned in my thesis, to achieve a score of 90\% on \textit{CIRAR10} dataset with the lowest number of CPU consumed. We compared our result with the CPU time required by DenseNet\cite{densenet} architecture to achieve the same accuracy.  The steps are as follows: we trained the model from Table \ref{tab:mymodel}, then retrained the last few layers and gathered six submodels using \textit{FGE} cyclic learning rate. Then, we retrained the last layer of each submodel once again, and eventually, we ensembled them into \textit{SuperNet}. Table \ref{tab:best_vs_densenet} shows CPU time required to do all the actions, with comparison to \textit{DenseNet}'s result. By using \textit{FGE} we can get 90\%+ accuracy much faster but we need to keep in mind that \textit{DenseNet} has a few times fewer parameters than \textit{SuperNet} which may impact prediction time. However, we have to remember also that our NN ensamble can be compresed substantially, according to \cite{ticket}.

\begin{table}[]
\centering
\begin{tabular}{|l|l|l|l|l|l|l|l|}
\hline
                                                                       & \textit{1'} & \textit{2'} & \textit{3'} & \textit{4'} & \textit{5'} & \textit{SuperNet} & \textit{Softmax voting} \\ \hline
\textit{FGE 1}                                                         & 0.8506      & 0.8347      & 0.825       & -           & -           & \textbf{0.877}    & 0.855            \\ \hline
\textit{FGE 2}                                                         & 0.8547      & 0.8601      & 0.8764      & 0.862       & -           & \textbf{0.8973}   & 0.8774           \\ \hline
\textit{FGE 3}                                                         & 0.8347      & 0.8195      & 0.8548      & 0.8211      & 0.825       & \textbf{0.8930}   & 0.8534           \\ \hline
\textit{\begin{tabular}[c]{@{}l@{}}Independent \\ models\end{tabular}} & 0.8619      & 0.8548      & 0.8434      & -           & -           & \textbf{0.9030}   & 0.8804           \\ \hline
\end{tabular}

\caption[\textit{FGE} vs independent models]{Comparison of validation accuracy  on \textit{CIFAR10} dataset, achieved by three different groups of models (Table \ref{tab:mymodel}) obtained with the \textit{FGE} method. As we can see, independent models reach a bit higher score with \textit{Supermodeling} approach, despite the lower number of submodels.}
\label{tab:fge}
\end{table}

\begin{table}[]
\centering
\begin{tabular}{|l|l|l|}
\hline
           & accuracy & time     \\ \hline
\textit{DenseNet}   & 0.9056   & 03:16:00    \\ \hline
\textit{FGE + SuperModel} & 0.9042   & 02:23:00 \\ \hline
\end{tabular}
\caption[\textit{FGE} as quick accuracy boost]{Validation accuracy on \textit{CIFAR10} dataset. Second row indicates a custom model (Table \ref{tab:mymodel}) on which we additionally trained the last layers. Then, we used \textit{FGE} technique to obtain 6 different submodels, and we trained their last layers once again. Eventually,  we build the \textit{Supermodel}. The time is the total CPU used for all those steps.}
\label{tab:best_vs_densenet}
\end{table}

\section{Conclusions and Future Work} 
\label{chap:conclusions-future}

\subsection{Summary and conclusions}

In our thesis, we developed and studied a specific kind of ensembling method of neural networks. In the chapter "Network Partitioning", we introduced the last layer training approach for individual networks. Then, we brought that concept to ensemble learning; we were merging multiple homogeneous dense models with the softmax layer, which we additionally fit for a short time. We measured the properties of predictions of such ensemble and compared it to the single networks' performance. In the chapter "Supermodeling of convolutional networks", we applied the same method for deeper and more complex architectures. We gathered good results for standard benchmarking datasets. Eventually, the chapter "Snapshots ensembles" tried to answer the question of whether we could obtain sub-models for an ensemble by taking snapshots of single network training with a specific learning rate.

Our thesis has a thoughtful structure; we started from fully-connected network partitioning and finished on deep architectures. It turns out that \textit{Supermodeling} technique seems to work in all the spectrum of models' complexity and size. By that, we wanted to show that there is a blurred boundary between considering an ensemble as a group of entirely independent models and a great, multi-component single learner. Following this lead, a single network is an ensemble of its neurons. Therefore, can we treat neurons and complete models as generic learning units, instead of identifying them separately? Can we regulate the correlation between such units in order to achieve better performance overall?  In the chapter "Network Partitioning", we have concluded that \textit{Supermodeling} might work as regularization for very shallow networks and achieves slightly better results than \textit{dropout}. The newest achievements in the domain\cite{wide_res_nets},\cite{stochastic} presumes that shallow networks solve the problem of \textit{"vanishing gradient"} and \textit{"diminishing feature reuse"} problems. For instance, \textit{WideResNet}\cite{wide_res_nets} architecture proposes shallow networks with usage of \textit{dropout} in the dense parts, which significantly helps the Authors in reducing overfitting. Therefore, can we replace \textit{dropout} with a method that builds less correlated units (paths) in such parts?
The answers for all the questions we leave to the readers for further studies.

\subsection{Future Work}

There are many exciting directions for further research. Generally, we would divide it into three major groups:

\begin{enumerate}
\item Last layer training - Why does it work? Does it benefit from some properties of \textit{softmax} loss functions? How portable is this method for different architectures and problems (not only classification tasks)? Can be this method improved so that it could be just the part of regular training and not the last stage? What are the disadvantages and bottlenecks?
We want to mention that our study around layer-by-layer training were independent and not inspired by the existing method called FreezeOut\cite{freezeout}, which we discovered in the latter stages of writing this thesis.
\item How submodels' correlation impacts ensemble's performance - We would continue the study around replacing \textit{dropout} with ensembles of different levels of correlation. How do we measure such a correlation? Is it possible to develop a new method that is better than \textit{dropout} and generic enough? Where is the sweet spot between sharing information between models and keeping them independent? Can we parameterize such an ensemble approach, similarly to the \textit{dropout} probability rate?
\item Ensemble's models' diversity - one can go through the literature that touches the topic of measuring the diversity of the models\cite{diversity}. How does it impact the final ensembles prediction? Are we able to generate diverse models (similarly to \textit{FGE}\cite{garipov} method) at even lower cost? How are diverse models located on the space of the loss function?
\end{enumerate}

Nevertheless, I am glad that we had this occasion to have this deep dive into the neural network's world. Hopefully, it is not the last time.

\bibliographystyle{ieeetr}

\addcontentsline{toc}{section}{\bibname}
\bibliography{thesis}
\printindex

\end{document}